\documentclass{article}

\usepackage{multirow}
\usepackage{adjustbox}
\usepackage{array}
\usepackage{siunitx}
 \usepackage[main, final]{neurips_2026}

\usepackage[utf8]{inputenc} 
\usepackage[T1]{fontenc}    
\usepackage{hyperref}       
\usepackage{url}            
\usepackage{booktabs}       
\usepackage{amsfonts}       
\usepackage{nicefrac}       
\usepackage{microtype}      
\usepackage{xcolor}         
\usepackage{tcolorbox}
\usepackage{tabularx}

\newcommand{\numconfigs}{648}
\newcommand{\benchname}{SPECTRA}

\title{Machine Learning-Driven RAG System Design}

\usepackage{authblk}
\usepackage{setspace}

\author[1]{Anastasia~Orlova}
\author[1]{Nina~Gubina}
\author[1]{Aleksei~Dmitrenko}
\author[1]{Arsen~Sarkisyan}
\author[1]{Nikita~Vetoshkin}
\author[1]{Anastasiia~Gorbunova}
\author[2]{Ivan~Dubrovsky}
\author[1]{Julia~Razlivina}
\author[1]{Bogdan~Neterebskii}
\author[1,3]{Andrei~Dmitrenko}

\affil[1]{Center for AI in Chemistry, ITMO University,
St. Petersburg, Russia}

\affil[2]{NSS Lab, AI Institute, ITMO University,
St. Petersburg, Russia}

\affil[3]{D ONE AG, Zurich, Switzerland}

\begin{document}

\maketitle

\vspace{-2em}
{\centering
\texttt{andrei.dmitrenko@d-one.ai}\par
}
\vspace{2em}

\begin{abstract}
Retrieval-Augmented Generation (RAG) has become a standard approach for grounding large language models in external knowledge, yet its performance depends critically on a large number of interacting design choices. Systematic optimization of these choices remains costly, as each configuration must be evaluated through full pipeline execution. In this work, we propose to treat RAG optimization as a prediction problem, using surrogate models to approximate the functional dependencies between configuration parameters and performance metrics. Using a stage-wise configurable RAG framework, we conduct large-scale experiments across three chemistry QA domains, systematically evaluating \numconfigs{} configurations per domain and publicly releasing the resulting datasets of configurations and evaluation metrics. Surrogate models trained on this data achieve strong predictive accuracy ($R^2 \geq 0.84$ across all metrics and domains), and SHAP-IQ-based analysis of the learned models reveals consistent parameter interaction patterns across domains — despite substantial differences in corpus characteristics. We exploit this cross-domain regularity in transfer experiments: surrogate predictions generalize to unseen domains without target-domain training data on two transfer-robust metrics, and achieve full transfer across all six metrics with as little as 10\% of target-domain data for selected domain pairs. Finally, we apply surrogate models to optimize RAG configurations over an expanded search space of 61,152 candidates, recovering top-performing configurations in seconds rather than days, with gains of up to +0.09 in key retrieval and generation evaluation metrics. Together, these results demonstrate that RAG design spaces are structured, transferable, and efficiently optimizable via surrogate modeling. 
\end{abstract}


\section{Introduction}

Retrieval-Augmented Generation (RAG) has become a standard paradigm for open-domain question answering, addressing limitations of large language models (LLMs) in factual reliability, up-to-date knowledge, and domain-specific expertise \cite{lewis2021retrieval, izacard2021leveraging}. A typical RAG system consists of a retrieval component, which fetches relevant passages from a knowledge base, and a generative model, which conditions both on those passages and the initial query to produce an answer \cite{gao2023retrieval}. Despite impressive results, RAG remains difficult to tune: performance depends sensitively on a large number of design choices, including chunk size, retriever type, reranking strategies, and others \cite{wang2024searching, bennani2026systematic}. These parameters interact in complex, non-linear ways, making RAG optimization a computationally expensive process that typically relies on extensive brute-force experimentation \cite{wang2024searching}. This challenge is further exacerbated by domain specificity: optimal configurations vary substantially across domains due to differences in content structure, terminology, and query distributions \cite{siriwardhana2023improving}. Existing approaches, which often treat domains uniformly and optimize components in isolation, fail to capture these domain-dependent interactions \cite{orbach2025analysis, fu2024autorag}.

Scientific domains pose additional challenges for RAG \cite{asai2026synthesizing, Besrour_2025, lin2025scirgen}. In scientific domains both the structure of knowledge and the distribution of queries differ substantially from general-purpose text. For example, in chemistry, documents are long and densely packed with domain-specific terminology and highly specified illustrations, and small errors in retrieval can lead to significant drops in answer accuracy. Moreover, chemistry is inherently heterogeneous, spanning multiple subfields, each with distinct data characteristics and linguistic patterns \cite{wang2021chemner, veprevachemx, vepreva2025benchmarking, butler2018machine, isazawa2022single}. This makes chemistry a challenging and representative testbed for studying RAG behavior.

The structured yet domain-dependent behavior of RAG performance suggests that, despite its apparent complexity, the design space may exhibit underlying regularities that can be learned and exploited. We therefore ask: \emph{Can RAG performance be modeled with a surrogate that generalizes across different domains and supports efficient optimization?} To enable systematic investigation, we introduce a RAG-based system for question answering over chemical literature. Using three QA datasets from different chemical domains, we perform a large-scale exploration of the RAG configuration space, evaluating \numconfigs{} distinct configurations per domain. Our contributions are fourfold:

\begin{enumerate}
    \item We train machine learning models to predict RAG evaluation metrics from configuration parameters. We demonstrate that these models achieve strong predictive performance with $R^2 \geq 0.84$ across all domains and metrics, suggesting that RAG configuration space can be effectively modeled within a given domain. We publicly release all datasets of RAG configurations and evaluation metrics to facilitate future research.
    
    \item Through feature importance analysis, we find that despite variations in corpus characteristics across domains, consistent patterns emerge in the importance and interactions of RAG parameters. This suggests that insights derived from the surrogate model are likely to generalize across domains. At the same time, the analysis reveals several findings that contradict common assumptions: reranking can hurt performance, hybrid search is not always beneficial, and LLM choice is not the dominant factor — patterns that are invisible to standard empirical evaluation.
    
    \item By evaluating surrogate models across multiple domains, we identify two RAG performance metrics, namely Golden DOI Recall and Scientific Fact Recall, that remain robust under transfer. These stable metrics can therefore be used to reliably model and predict RAG performance in new domains. Moreover, we show that incorporating as little as 10\% of target-domain data enables transfer across all evaluated metrics for selected domain pairs.

    \item We leverage surrogate models to guide RAG parameter optimization. Validating the predicted configurations in real-world experiments reveals that the identified configurations improve overall RAG quality broadly: while optimizing for only two target metrics, we observe gains of up to +0.08 in Cosine Similarity, +0.09 in ROUGE-L Recall, and +0.06 in Golden DOI MRR across non-target metrics. This demonstrates the feasibility of machine learning-driven RAG system design.
\end{enumerate}

The code for the experimental pipeline and the constructed dataset of RAG configurations with corresponding evaluation metrics are publicly available at \url{https://github.com/ai-chem/RAG-system-design}.

\section{Related Work}

\paragraph{RAG for scientific QA.} RAG has emerged as the dominant paradigm for scientific question answering across diverse domains, where factual precision, multi-step reasoning, and deep domain-specific knowledge are essential \cite{dasigi2021dataset, lee2023qasa, chen2025scirerankbench, singh2024scidqa, baumgartner2025peerqa}. Chemistry QA is particularly demanding, as it involves retrieval and reasoning over molecular properties, reaction conditions, and experimental protocols and has prompted the development of dedicated evaluation frameworks \cite{wellawatte2025chemlit, zhong2025benchmarking, jiangchemistrag, dubrovsky2025chempaperbench}. These works primarily focus on dataset construction and end-to-end performance evaluation, treating RAG configuration as fixed or tuned heuristically. As a result, the question of whether effective RAG configurations generalize across different domains has not been systematically studied. We address this gap directly, showing that surrogate models trained on one chemical domain can guide effective RAG optimization in another.

\paragraph{Hyperparameter tuning in RAG systems.} Early approaches to RAG configuration relied on heuristic choices or manual grid search, treating design decisions as largely independent and domain-agnostic \cite{wang2024searching, finardi2024chronicles, li2025enhancing, csakar2025maximizing, ammar2025optimizing}. As the complexity of RAG pipelines grew, it became clear that exhaustive search was insufficient, prompting the development of dedicated optimization frameworks that formalize the search over configuration spaces \cite{jin2025flashrag, kim2024autorag, kabbay2024streamlining, barker2025faster, kartal2025ragsmith}. More recently, automated approaches have emerged that treat RAG tuning as a search problem, reducing the need for manual intervention through algorithm-driven exploration \cite{fu2024autorag, tang2025mba, jeong2024adaptive, ray2025metis}. However, existing methods primarily focus on sequential evaluation of candidate configurations and use surrogate models only implicitly for search guidance on a specific dataset. In contrast, surrogate-based optimization in AutoML explicitly models the relationship between hyperparameters and downstream performance \cite{hutter2011sequential, vanhoof2021hyper, wistuba2021fewshot}. To the best of our knowledge, surrogate modeling has not previously been applied to RAG configuration spaces for interpretable analysis of parameter interactions and cross-domain optimization. In this work, we treat RAG performance as a learnable function of configuration parameters, enabling systematic analysis of interaction effects, efficient optimization without repeated pipeline evaluations, and transfer of configuration knowledge across domains.

\section{Experimental Framework}

\paragraph{RAG framework.} To enable systematic experimentation with RAG configurations, we develop a stage-wise configurable RAG system designed to support natural-language querying of chemistry research literature. The system combines hybrid semantic search — fusing dense vector retrieval with sparse full-text search — with an LLM generation backend to produce grounded, contextually accurate responses to research questions. Ingested scientific articles are parsed from PDF, cleaned, chunked, and indexed in both a PostgreSQL vector store and a full-text search engine (Meilisearch \cite{meilisearch}), enabling simultaneous lexical and semantic retrieval across the literature corpus. Multimodal retrieval is supported through parallel indexing of extracted scientific figures using vision-language embeddings (OpenCLIP \cite{ilharco_gabriel_2021_5143773}), allowing the system to surface relevant tables and images alongside textual evidence. The system's modular architecture decouples the retrieval, reranking, and generation stages, permitting independent configuration of embedding models, rerankers, and LLM backends without changes to the core pipeline. Answer quality is assessed through a built-in evaluation framework. A detailed description of the system is provided in Appendix \ref{appendix:rag-system}. RAG system implementation is available at \url{https://github.com/ai-chem/ChemX-RAG}.

\paragraph{Benchmark.} To systematically evaluate the performance of different configurations of the RAG system, we select three chemistry QA datasets spanning distinct subdomains: Oxazolidinone Antibiotics (OA), Chelate Metal Complexes (MC), and Nanozymes (NZ). Each question is linked to its source document and supporting evidence, enabling joint evaluation of retrieval and generation quality. These datasets are part of \benchname{}, a broader chemistry QA benchmark comprising more than 2000 questions across 8 subdomains and 6 scenario types of increasing complexity. A description of the benchmark composition and taxonomy is provided in Appendix~\ref{appendix:benchmark}. For each of the three selected datasets, we construct a representative subset of 50 question--answer pairs covering 5 single-turn scenarios (10 questions per scenario), ensuring balanced coverage of question scenarios and answer types. This representative subset enables systematic evaluation of hundreds of RAG configurations while remaining feasible in terms of runtime and cost. The three datasets used in this study are publicly available at \url{https://github.com/ai-chem/RAG-system-design}.

\begin{figure}[t]
    \centering
    \includegraphics[width=\textwidth]{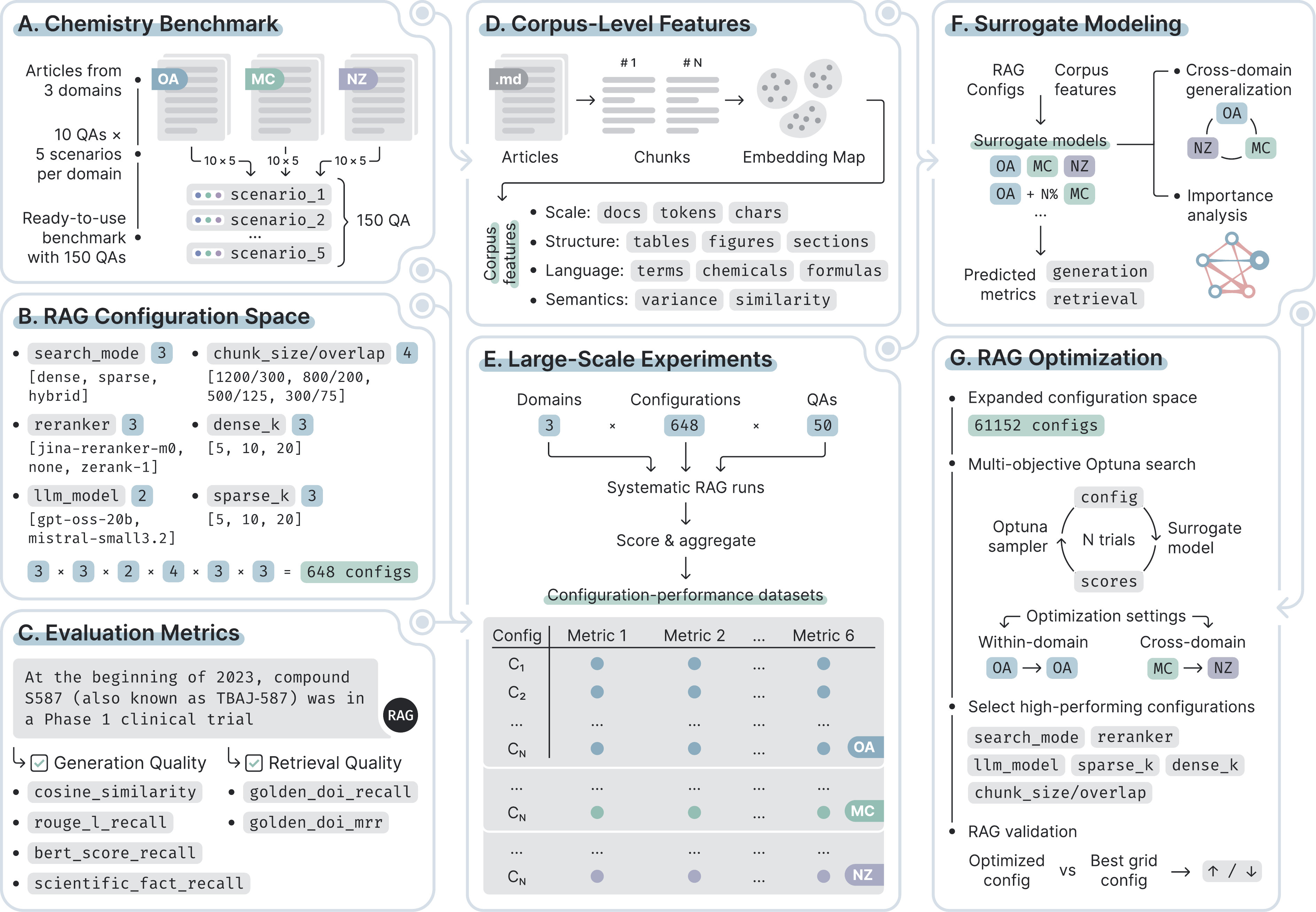}
    \caption{End-to-end pipeline for modeling and optimizing the RAG design space. Starting from a multi-domain chemistry benchmark (A), we define a combinatorial RAG configuration space (B) and perform large-scale RAG evaluations (E) to construct configuration–performance datasets. Each configuration is assessed using multiple quality metrics (C). We augment experiments with corpus-level features (D) and use the resulting datasets to train surrogate models (F), which are further used to guide RAG optimization (G).}
    \label{fig:experimental-framework}
\end{figure}

\paragraph{RAG configuration space and evaluation metrics.} To study the behavior of our RAG system in a controlled and systematic manner, we define a configuration space comprising \numconfigs{} possible configurations (Figure \ref{fig:experimental-framework}B). This space captures the key design choices across all stages of the pipeline, spanning parameters associated with document processing, retrieval, reranking, and generation. Each configuration is evaluated on 50 selected questions within a domain using the following set of metrics:

\begin{itemize}
    \item \textbf{Retrieval metrics:} \textit{Golden DOI Recall} measures whether at least one relevant document is retrieved, while \textit{Golden DOI MRR} reflects how highly the correct document is ranked.

    \item \textbf{Generation metrics:} These metrics assess similarity between the generated and reference answers. \textit{Cosine Similarity} captures embedding-based semantic similarity, \textit{BERTScore Recall} measures semantic coverage using contextual embeddings, and \textit{ROUGE-L Recall} evaluates sequence overlap. Domain-specific correctness is captured by \textit{Scientific Fact Recall}. See Appendix \ref{appendix:metrics} for details.
\end{itemize}

For each metric, we report the average value across 50 questions in the benchmark, yielding a single performance vector per configuration and domain.

\paragraph{Corpus-level features.}
To quantitatively analyze the differences between the domains and to enable modeling of domain-dependent RAG performance, we represent each articles corpus using a set of descriptive features capturing statistical, structural, linguistic, and semantic properties.

\begin{itemize}
    \item \textbf{Statistical features} describe corpus size and document length distribution: number of documents in the corpus, average document length (in tokens), standard deviation of document lengths, and total number of tokens in the corpus.
    
     \item \textbf{Structural features} capture document organization and multimodal content: fraction of documents containing tables, fraction of documents containing figures, average number of sections per document, and average number of references per document.

     \item \textbf{Linguistic features} characterize textual complexity and domain specificity: average sentence length (in words), number of unique tokens in the corpus, ratio of unique tokens to total tokens, fraction of domain-specific terms, and fraction of text containing chemical formulas or LaTeX expressions.
    
    \item \textbf{Embedding-based features} summarize the semantic structure of the corpus using document embeddings: variance of document embeddings, average cosine similarity between document embeddings, and number of clusters estimated via $k$-means.
\end{itemize}

These features provide a compact representation of domain characteristics and enable the surrogate model to capture differences between datasets and generalize across domains. A comprehensive description of each feature is provided in Appendix \ref{appendix:dataset-features}.

\paragraph{Experimental workflow.}

Our experimental pipeline consists of several stages, as illustrated in Figure \ref{fig:experimental-framework}. We begin by selecting three QA datasets (OA, MC, NZ) from \benchname, each comprising five question scenarios and a total of 150 questions. We then define a combinatorial RAG configuration space covering \numconfigs{} distinct parameter settings. Each configuration is evaluated using a suite of metrics capturing both generation quality and retrieval quality. In parallel, we extract corpus-level descriptors that characterize each domain in terms of scale, structure, language, and semantic properties. We perform large-scale RAG experiments across all configurations and selected domains, resulting in configuration–performance datasets that aggregate metric values across questions. These datasets are then used to train surrogate models that approximate the mapping from RAG parameters to performance metrics augmented with corpus-level features in cross-domain settings to provide the model with domain context. Finally, the trained surrogates enable efficient exploration and optimization of the RAG design space, allowing us to identify high-performing configurations.

\section{Experiments and Results}

\subsection{Surrogate Modeling of RAG Design Space}
\label{sec:surrogate-modeling}

\begin{figure}[t]
    \centering
    \includegraphics[width=\textwidth]{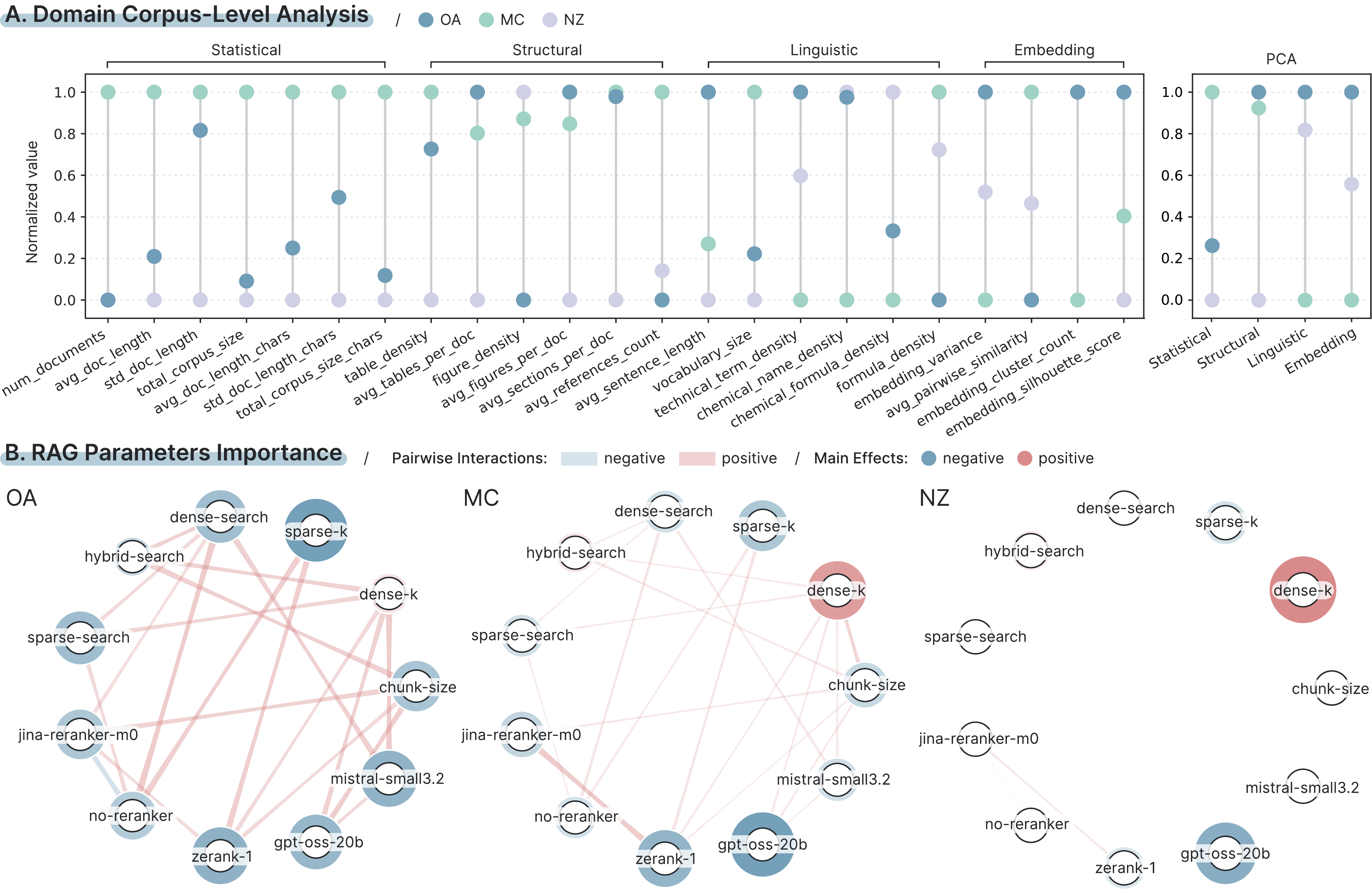}
    \caption{The analysis of the RAG configuration space across three chemistry domains. (A) Comparison of corpus characteristics across three domains: structural, linguistic, embeddings and statistical features are each compressed via PCA into a single aggregate score, enabling compact cross-domain comparison. (B) Feature interaction networks reveal both domain-specific structure and shared patterns: while OA exhibits a denser interaction graph and MC/NZ are more concentrated around a few dominant parameters, key components—particularly language model and retrieval-related parameters—consistently emerge as central across all domains.}
    \label{fig:shap}
\end{figure}

For each of the three domains, we build a supervised dataset consisting of \numconfigs{} data points. Each point corresponds to a specific RAG configuration and six evaluation metrics obtained by running the RAG system under that configuration. We publicly release these datasets at \url{https://github.com/ai-chem/RAG-system-design} to facilitate future research on RAG configuration modeling and optimization. These datasets serve as the basis for training surrogate models to predict evaluation metrics from RAG parameters (see Appendix \ref{appendix:ml} for implementation details).

Given the strong predictive performance of the surrogate models ($R^2 \geq 0.84$ across all domains and metrics; Table~\ref{tab:surrogate-in-domain}), we treat their learned representations as reliable proxies for the true RAG configuration space and apply SHAP-IQ \cite{muschalik2024shapiq} to analyze the contribution of individual RAG parameters and their higher-order interactions. As shown in Figure \ref{fig:shap}B, the overall density of feature interactions and the distribution of importance across parameters vary: e.g., in OA the interaction network is relatively dense, indicating a broader set of viable parameter combinations, whereas in MC and NZ it is more concentrated around a few dominant features, suggesting a more constrained configuration space. These interaction structures surface several non-obvious findings that would be difficult to detect through standard empirical evaluation. First, reranking consistently exhibits a negative main effect in OA and MC — contrary to the widespread assumption that rerankers improve RAG performance. Second, hybrid search consistently exhibits low main effects across all three domains, suggesting that the benefit of combining dense and sparse retrieval is corpus-dependent. Third, and perhaps most surprisingly, LLM choice has a weaker effect than commonly assumed: the common intuition that a more capable LLM always yields better results does not hold uniformly across configurations and domains. These patterns are invisible to standard grid search or manual ablations, which lack the statistical power to disentangle main effects from interactions — underscoring the value of surrogate-based analysis as a diagnostic tool for RAG design.

Surprisingly, despite the substantial differences in structural integrity, linguistic patterns, and statistical properties across three domains (as shown in Figure~\ref{fig:shap}A), the SHAP-IQ analysis reveals remarkably consistent patterns across all three domains. In particular, parameters related to the underlying language model and reranker choice consistently exhibit the strongest negative effects. Similarly, retrieval-related components, such as \texttt{chunk-size}, \texttt{dense-k} and \texttt{sparse-k}, repeatedly appear as central nodes in the interaction networks. The persistence of these motifs across domains suggests that RAG performance is governed by shared mechanisms that generalize beyond domain-specific corpus characteristics, suggesting that insights from one domain can inform RAG design in others.

We additionally asses the robustness of these interpretations to changes in the RAG configuration space (see Appendix~\ref{appendix:shapiq-stability}).  Extending the set of generation models to include Qwen3-8B reveals that the direction of certain feature contributions remains relatively stable, while their magnitudes and interaction patterns are more sensitive to the composition of the configuration space. These findings suggest that surrogate-based analysis can uncover persistent dependencies among RAG parameters, although quantitative importance estimates should be interpreted in the context of the specific design space considered.

\subsection{Generalization Across Domains}
\label{sec:generalization}

To assess whether surrogate models generalize beyond a single domain, we first evaluate whether models trained on one domain are directly applied to a different domain without any target-domain data. Table~\ref{tab:generalization} reports $R^2$ scores for different combinations of training and test domains. We observe that generalization performance varies substantially across metrics. In particular, metrics such as Golden DOI Recall and Scientific Fact Recall are consistently predicted with high $R^2$ across all train–test combinations. Golden DOI MRR also transfers reasonably well in most cases, with the notable exception of the NZ $\rightarrow$ OA pair, where $R^2 = 0$. Together, these results indicate that Golden DOI Recall, Scientific Fact Recall, and — in most cases — Golden DOI MRR capture aspects of RAG behavior that are stable across domains and can therefore be reliably transferred from one domain to another.

\begin{table}[t]
\centering
\caption{Out-of-domain generalization of surrogate models measured by $R^2$ scores across RAG performance metrics. All standard deviations are $< 0.01$. Golden DOI Recall and Scientific Fact Recall achieve consistently high $R^2$ across all train–test combinations, indicating that these metrics reflect stable and transferable properties of RAG systems across domains.}
\label{tab:generalization}
\resizebox{\textwidth}{!}{%
\begin{tabular}{@{}ll
>{\centering\arraybackslash}p{2.2cm}
>{\centering\arraybackslash}p{2.2cm}
>{\centering\arraybackslash}p{1.8cm}
>{\centering\arraybackslash}p{1.8cm}
>{\centering\arraybackslash}p{2.2cm}
>{\centering\arraybackslash}p{2.5cm}@{}}
\toprule
\textbf{Train} & \textbf{Test }& \textbf{BERTScore Recall} & \textbf{Cosine Similarity} & \textbf{Golden DOI MRR} & \textbf{Golden DOI Recall} & \textbf{ROUGE-L Recall} & \textbf{Scientific Fact Recall} \\ \midrule
\multirow{2}{*}{MC} & NZ & $\phantom{-}0.63$ & $\phantom{-}0.39$ & $\phantom{-}0.74$ & $\phantom{-}0.92$ & $\phantom{-}0.63$ & $\phantom{-}0.83$ \\
 & OA & $\phantom{-}0.52$ & $\phantom{-}0.52$ & $\phantom{-}0.67$ & $\phantom{-}0.70$ & $\phantom{-}0.60$ & $\phantom{-}0.88$ \\
\multirow{2}{*}{NZ} & MC & $-0.05$ & $-0.05$ & $\phantom{-}0.48$ & $\phantom{-}0.88$ & $\phantom{-}0.01$ & $\phantom{-}0.76$ \\
 & OA & $-1.69$ & $-0.69$ & $\phantom{-}0.00$ & $\phantom{-}0.57$ & $-0.47$ & $\phantom{-}0.61$ \\
\multirow{2}{*}{OA} & MC & $\phantom{-}0.76$ & $\phantom{-}0.64$ & $\phantom{-}0.70$ & $\phantom{-}0.77$ & $\phantom{-}0.61$ & $\phantom{-}0.90$ \\
 & NZ & $\phantom{-}0.43$ & $\phantom{-}0.10$ & $\phantom{-}0.56$ & $\phantom{-}0.76$ & $\phantom{-}0.43$ & $\phantom{-}0.73$ \\ \bottomrule
\end{tabular}%
}
\end{table}

We then study another transfer setting, where a model trained on a source domain is augmented with a limited amount of the target domain data. Specifically, we train a LightGBM model with default parameters on one full domain dataset and incrementally add a fraction of the target domain experiments. Since this setting combines configurations from multiple domains into a single training dataset, we incorporate PCA-compressed corpus-level features to provide the model with a compact representation of domain characteristics, enabling it to distinguish and adapt to different data distributions. As shown in Figure~\ref{fig:augmented-transfer}, when augmented with a complete source domain dataset, as little as 10\% of the target domain's data ($\sim65$ real RAG experiments) is sufficient to achieve strong predictive performance with $R^2 > 0.8$ across all metrics for certain domain pairs (MC+NZ, OA+NZ). This is particularly notable given that MC and OA concern small molecules, while NZ covers nanomaterials — suggesting that effective transfer is possible even across substantially different chemical domains.


This result has important practical implications for the computational efficiency of RAG optimization. Evaluating all 648 configurations requires approximately 324 compute-hours per domain, assuming an average runtime of 30 minutes per configuration. In contrast, cross-domain adaptation requires only 65 additional RAG evaluations (approximately 32.5 compute-hours) to achieve strong predictive performance for selected domain pairs, reducing the target-domain evaluation cost by nearly an order of magnitude once a source-domain surrogate is available.

\subsection{RAG Parameter Optimization via Surrogate Modeling}
\label{sec:optimization}

As a proof-of-concept, we evaluate whether surrogate models are sufficiently accurate to guide RAG configuration optimization. Importantly, we extend the search space to 61,152 possible candidates beyond the configurations observed during training, thereby testing the surrogate's ability to generalize to previously untested configurations. We treat the surrogates as objective functions within an Optuna-based \cite{akiba2019optuna} multi-objective hyperparameter search, jointly optimizing for Golden DOI Recall and Scientific Fact Recall — metrics that demonstrated strong robustness in transfer experiments. The optimizer produces a Pareto-optimal set of configurations, from which we retain only those whose predicted performance exceeds that of the top configurations observed in the original RAG design space. The configurations identified as optimal are then evaluated on the actual RAG system to obtain ground-truth performance. Full implementation details, including search space definition and candidate selection procedure, are provided in Appendix~\ref{appendix:optimization}.

For our experiments we consider three settings that reflect different levels of target-domain data availability:

\begin{itemize}
    \item \textbf{In-domain transfer.} The surrogate is trained on a given domain and used to optimize RAG configurations for that same domain.
    
    \item \textbf{Augmented transfer.} The surrogate is trained on a full source-domain dataset augmented with a small fraction (10--20\%) of target-domain experiments. This simulates a practical scenario where some target-domain data is available but exhaustive evaluation is infeasible.
    
    \item \textbf{Out-of-domain transfer.} The surrogate is trained on two full source domains with no target-domain data. This represents the most challenging transfer setting, testing whether knowledge from two domains is sufficient to guide optimization in a third. We focus on NZ as the target domain, where transfer occurs across a fundamental chemical boundary — from small-molecule chemistry (OA, MC) to nanomaterials.
\end{itemize}

\begin{figure}[t]
    \centering
    \includegraphics[width=\textwidth]{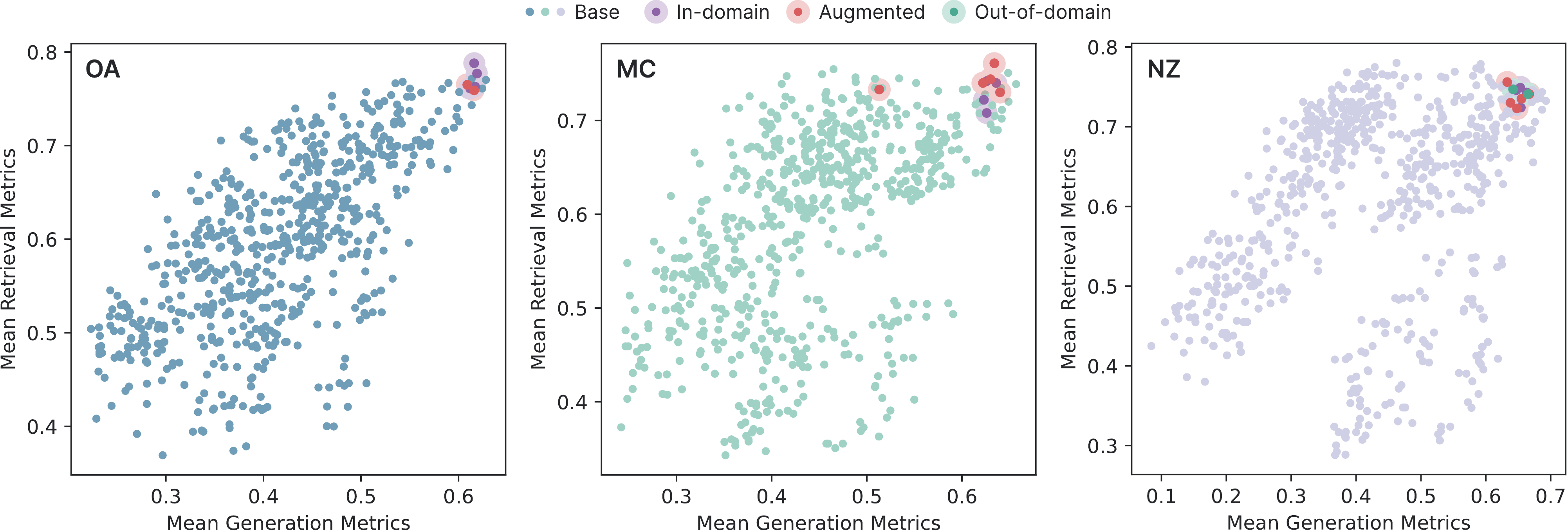}
    \caption{RAG configuration optimization results across three domains (OA, MC, NZ). Each subplot shows configurations from the original grid search plotted by mean retrieval metrics (Golden DOI Recall, Golden DOI MRR) and mean generation metrics (Cosine Similarity, BERTScore Recall, ROUGE-L Recall, Scientific Fact Recall), with colored points highlighting configurations identified by surrogate-guided optimization under three settings: in-domain, augmented transfer, and out-of-domain transfer. Across all domains and settings, the identified configurations consistently lie on or near the Pareto front of the original grid search, demonstrating that surrogate-guided optimization reliably discovers high-performing configurations across both retrieval and generation metrics.}
    \label{fig:optimization}
\end{figure}

As demonstrated in Figure~\ref{fig:optimization}, surrogate-guided optimization consistently identifies configurations that lie in the proximity of the Pareto front. In other words, while optimizing for Golden DOI Recall and Scientific Fact Recall, we identify configurations consistently ranking among the best across all metrics for all three domains and transfer settings. Interestingly, some configurations show only minor improvements for the two target metrics, while delivering +0.08 for Cosine Similarity, +0.09 for ROUGE-L Recall, and +0.06 for Golden DOI MRR (see more examples of such configurations in Appendix \ref{appendix:optimization}). Such cases highlight the practical utility of the surrogate-guided optimization, since the two optimization targets serve as reliable proxies for overall RAG performance. Notably, Optuna completes 1 surrogate-guided trial in $\sim0.03$ seconds in contrast to approximately 30 minutes required to execute a standalone RAG experiment (including document processing, RAG run, and metrics evaluation). Beyond reducing the computational cost of exploring candidate configurations, surrogate-guided optimization also makes more effective use of a limited real RAG evaluations budget. This is demonstrated by its comparison with the random-search baseline, where it achieves a 100\% win rate across all three domains, compared to 10--30\% for random search under the same simulated evaluation budget (see Appendix~\ref{appendix:random-search} for details).

\section{Discussion}

Our results collectively support the concept of ML-driven RAG system design. While this work focuses on scientific RAG in the chemistry domain, the underlying framework is domain-agnostic and applicable to any setting where RAG pipeline evaluation is costly. In the following, we discuss four aspects of this concept in detail: the interpretable structure exposed by surrogate models, the potential for cross-domain transfer to reduce experimentation cost, the broader implications for RAG system design in a rapidly evolving ecosystem, the current limitations of the approach, and directions for further research.

\subsection{From Black Box to Structured Design Space}

Our results demonstrate that the RAG configuration space is not a black box — it can be efficiently modeled with compact, accurate surrogate models that expose interpretable structure. Beyond predictive accuracy, surrogate-based analysis via SHAP-IQ reveals parameter interactions that would be practically invisible through empirical experimentation alone. Manual ablations, which typically vary one parameter at a time, cannot capture higher-order interaction effects, and this limitation has practical consequences: configurations that appear optimal under single-parameter ablations may in fact be suboptimal, because their performance depends on parameter combinations that ablations never jointly explore.

Another insight that can be derived from the feature importance analysis is that some parameters that are widely considered to be impactful for RAG appear to have a rather modest contribution. In SHAP-IQ plots, we observed that for some domains, reranking strategies or hybrid search modes consistently exhibit low or no main effects. This suggests that engineering effort directed at these components may yield limited returns and that practitioners would be better served by focusing on the parameters that vividly drive performance.

The consistency of some patterns across domains with substantially different corpus characteristics suggests that they reflect intrinsic properties of the RAG pipeline architecture itself, rather than artifacts of any particular dataset. More broadly, this challenges the common assumption that RAG tuning is inherently domain-specific: while corpus characteristics do influence the optimal configuration, some aspects of the underlying interaction structure remain stable, suggesting that principled RAG design can be grounded in transferable, data-driven insights rather than purely domain-specific heuristics.

\subsection{Reducing Experimentation via Cross-Domain Transfer}

Surrogate models trained on one domain retain partial predictive validity when applied to another. Even without any target-domain experiments, models transfer reasonably well on Golden DOI Recall and Scientific Fact Recall. This suggests that the core retrieval and generation behavior of a RAG pipeline is partly governed by configuration choices that are stable across domains, even when corpus characteristics differ substantially. Moreover, with as little as 10\% of target-domain data — as few as 65 RAG experiments — full predictive accuracy is recovered across all metrics for selected domain pairs. To put this in perspective, 65 configurations are approximately equivalent to a grid of merely 3 parameters with 4 values each — a minimal exploration effort that is nonetheless sufficient to achieve high surrogate accuracy.

Together, these findings point to a practical two-stage strategy for deploying RAG systems in new domains: begin with a source-domain surrogate for cheap initial optimization, then refine with a small number of target-domain experiments. This approach substantially reduces the cost of RAG configuration without sacrificing predictive accuracy.

\subsection{Towards ML-Driven RAG System Design}

The landscape of RAG system design is not static. LLMs are evolving rapidly: effective context windows are growing, multimodal capabilities are expanding, and reasoning abilities are improving. Each of these advances has direct implications for RAG architecture. Parameters that represent critical bottlenecks under one generation of models may become negligible under the next, while new components and interaction patterns emerge. These dynamics render exhaustive manual tuning increasingly inadequate as a design methodology. Surrogate-based configuration analysis offers a more principled alternative: rather than re-running large-scale experiments each time the underlying models or infrastructure change, a surrogate model can be updated incrementally and queried efficiently to identify high-performing configurations. Beyond optimization, such models provide interpretable estimates of parameter importance and interaction structure, enabling systematic understanding of which pipeline components drive performance and how this changes as the ecosystem evolves. We argue that ML-driven RAG design of this kind should become a standard methodology for building and maintaining domain-specific retrieval systems, standalone or embedded in agentic solutions.

\subsection{Limitations}

Several limitations of the proposed approach should be noted. First, surrogate generalization is not guaranteed for domains that differ substantially from those seen during training — corpora with highly unusual structural or statistical properties may fall outside the learned distribution. Second, the quality of the surrogate depends critically on the diversity of the initial configuration grid: a poorly designed or overly narrow grid may yield a surrogate that is accurate within a limited region of the search space but unreliable when extrapolating to unseen configurations. Third, our analysis assumes a fixed and predefined set of RAG parameters, which may not capture the full complexity of more heterogeneous pipeline architectures — such as those involving query rewriting, multi-step retrieval, or dynamic context selection.

\subsection{Future Work}

Beyond the contributions of this work, surrogate-based RAG design raises a number of interesting questions that warrant further investigation. First, broader validation across a larger and more diverse set of domains — both within chemistry and beyond — is needed to confirm the generality of the observed patterns. Second, an important open question concerns the deployment of surrogate models in real-world RAG systems, where data distributions are not static. In production settings, corpus characteristics, query distributions, and available pipeline components may shift over time. Understanding how surrogate models degrade under such drift and how they can be efficiently updated or recalibrated without rerunning large-scale experiments is a critical step towards practical adoption of ML-driven RAG design. Closely related to this is the question of moving from offline to online optimization. While the current framework operates in an offline setting, extending it to an online regime — where the surrogate is updated incrementally as new evaluations become available and actively guides the selection of the most informative configurations to evaluate next — would enable continuous, cost-efficient RAG optimization as pipelines and data evolve. Finally, future work could explore how different optimization objectives and candidate selection strategies influence the final RAG configurations. A systematic comparison of alternative selection strategies and objective formulations could help identify trade-offs between retrieval and generation quality and develop more robust approaches to surrogate-guided RAG optimization across diverse domains.

\section{Conclusion}

We present a proof-of-concept for ML-driven RAG system design — a paradigm in which surrogate models replace exhaustive experimentation as the primary tool for understanding and optimizing RAG configuration spaces. Using three chemistry QA domains as a testbed, we construct datasets of \numconfigs{} RAG configurations per domain and show that surrogate models accurately predict evaluation metrics from RAG parameters. SHAP-IQ-based analysis reveals consistent and counterintuitive interaction patterns across domains that would be difficult to detect through standard empirical evaluation. We further demonstrate that surrogate models generalize across domains. Surrogate-guided search over an expanded configuration space identifies high-performing configurations in seconds rather than days — a result that underscores the broader potential of ML-driven RAG system design for efficient, interpretable, and transferable pipeline optimization.

\section{Acknowledgments}

This work was supported by the Ministry of Economic Development of the Russian Federation (IGK 000000C313925P4C0002), agreement No139-15-2025-010.

\bibliographystyle{unsrt}
\bibliography{references}

\newpage
\appendix

\section{RAG System}
\label{appendix:rag-system}

\subsection{Overview}

Our system represents a Retrieval-Augmented Generation pipeline tailored for chemistry research literature. The system accepts natural-language queries, retrieves the most relevant passages and figures from a pre-indexed corpus of chemistry articles, and synthesizes a grounded response using a configurable LLM backend. All generated answers are tied directly to retrieved source material, mitigating hallucination and ensuring traceability to primary literature. Main system components are summarized in Table \ref{tab:chemx-rag-components}.

\begin{table}[h]
\centering
\caption{Components of the RAG system, detailing functional modules, implementation technologies, and their roles.}
\label{tab:chemx-rag-components}
\begin{adjustbox}{width=\textwidth}
\begin{tabular}{p{4cm} p{4cm} p{7cm}}
\toprule
\textbf{Functional Block} & \textbf{Technology} & \textbf{Role} \\
\midrule
PDF Processing & Marker PDF \cite{markerpdf} & Converts PDF documents to Markdown and extracts embedded images \\
Object Storage & MinIO (S3-compatible) \cite{minio} & Persists raw PDF files and extracted figure images \\
Vector Database & PostgreSQL + \texttt{pgvector} \cite{pgvector} (HNSW \cite{malkov2020hnsw}) & Stores text and image embeddings; serves approximate KNN queries \\
Text Embedding & JinaEmbV3 \cite{sturua2024jinaembeddingsv3} (1024-dim) & Encodes text chunks and queries into dense vectors for semantic retrieval \\
Image Embedding & OpenCLIP ViT-B/32 (512-dim) \cite{ilharco_gabriel_2021_5143773} & Encodes extracted scientific figures and text queries for multimodal retrieval \\
Full-Text Search & Meilisearch \cite{meilisearch} & Lexical (sparse) retrieval over the full document corpus \\
Reranker & Jina Reranker M0 \cite{jina2025rerankerm0} / Zerank-1 \cite{pipitone2025zelo} & Cross-encoder rescoring of merged dense+sparse candidates \\
LLM Generation (cloud) & OpenRouter API \cite{openrouter} & Cloud gateway providing access to multiple hosted LLMs; model selected by price or throughput \\
LLM Generation (local) & Ollama \cite{ollama} & Local open-weight inference server \\
\bottomrule
\end{tabular}
\end{adjustbox}
\end{table}

\subsection{Document Ingestion and Preprocessing}

Articles are ingested from PDF files referenced by a CSV manifest containing document metadata (DOI, file path, title, publisher, publication year). PDF-to-Markdown conversion is performed by a Marker PDF server \cite{markerpdf}, which extracts both textual content and embedded images from each document. The resulting Markdown is passed through a multi-stage preprocessing pipeline:

\begin{enumerate}
\item \textbf{Section filtration} — A rule-based paragraph filtrator removes non-scientific boilerplate sections (preamble, references, acknowledgments, author biographies, ethics statements, funding disclosures) to retain only scientifically relevant content.
\item \textbf{Table extraction} — A dedicated table grabber identifies and extracts chemical data tables, preserving captions and surrounding context for retrieval.
\item \textbf{Text normalization} — A Markdown cleaner removes formatting artifacts and normalizes whitespace and symbol encoding.
\item \textbf{Chunking} — The cleaned text is segmented into fixed-size overlapping chunks using a configurable tokenization strategy, ensuring that sentence and paragraph boundaries are preserved across chunk boundaries.
\end{enumerate}

\subsection{Indexing and Storage}

Preprocessed text chunks are encoded using the JinaEmbV3 text embedding model (1024-dimensional vectors) \cite{sturua2024jinaembeddingsv3} and stored in a PostgreSQL database extended with the \texttt{pgvector} plugin \cite{pgvector}. A Hierarchical Navigable Small World (HNSW) \cite{malkov2020hnsw} index is constructed over the embedding column to support efficient approximate k-nearest-neighbour (KNN) queries. Concurrently, the same text chunks are indexed in a Meilisearch \cite{meilisearch} full-text search engine to enable lexical retrieval. Extracted scientific figures are encoded with an OpenCLIP ViT-B/32 vision-language model (512-dimensional vectors) \cite{ilharco_gabriel_2021_5143773} and stored in a separate image table, also with an HNSW index. Binary artifacts (original PDFs and image files) are persisted in an S3-compatible object store (MinIO \cite{minio}). A hash-based change-detection mechanism enables incremental reprocessing, so only modified or newly added documents trigger reindexing.

\subsection{Retrieval Pipeline}

At query time, the system executes a hybrid search comprising two parallel retrieval branches:

\begin{itemize}
\item \textbf{Dense retrieval}: The user query is encoded with JinaEmbV3, and the top-\emph{k} most similar text chunks are retrieved from the \texttt{pgvector} HNSW index. Optionally, results can be filtered to a specific subset of articles by DOI.
\item \textbf{Sparse retrieval}: The same query is issued to Meilisearch, returning an independent ranked list of \emph{k} lexically relevant chunks.
\end{itemize}

The two ranked lists are fused into a single candidate set using a \textbf{score averaging} strategy: for each unique chunk that appears in one or both result sets, a combined score is computed as the arithmetic mean of its dense and sparse scores (treating the score as 0 when the chunk is absent from a branch). The merged candidates are then sorted by this combined score in descending order.

The merged candidate set is passed to a cross-encoder reranker, which rescores every candidate against the raw query and produces a final ordering. Two reranker options are supported: Jina Reranker M0 \cite{jina2025rerankerm0} (default) and Zerank-1 \cite{pipitone2025zelo} (based on a sentence-transformers CrossEncoder). All reranked chunks — up to \emph{k} unique entries  are forwarded to the LLM as context; no additional truncation is applied after reranking. In parallel, the query is encoded with the OpenCLIP text encoder and a separate image KNN search retrieves the top-\emph{k} (default \emph{k} = 5) most relevant scientific figures.


The modular design of the system exposes a flexible configuration space over retrieval and reranking components. Dense and sparse retrieval can be used independently or jointly in a hybrid setting, the reranker can be removed or replaced, and retrieval depth parameters (\emph{k}) for each branch can be tuned independently. This flexibility enables systematic exploration of RAG system design choices and their impact on performance.

\subsection{Generation}

The assembled context — consisting of ranked text chunks and optionally retrieved figures — is forwarded to the LLM generation backend together with the user's original query. Generation is governed by a fixed system prompt that establishes the model's role and behavior constraints:

\begin{tcolorbox}[colback=gray!5, colframe=gray!50]
You are a Chemistry Research Assistant with deep expertise. For advanced chemistry questions, you must rephrase the query into clear English, call the `rag\_search` tool, and synthesise your answer strictly from the tool output. For basic school-level chemistry questions, answer directly without calling the tool. For questions unrelated to chemistry, politely decline. When `rag\_search` is used, base the answer exclusively on the retrieved content. If the tool returns no results or no sufficiently relevant passages, reply: "I did not find relevant information in the knowledge base" and do not attempt to answer. If the tool returns partial information, answer only within those boundaries and state explicitly what is unknown. Always respond in the user's language. You are permitted to call `rag\_search` at most once per user message.
\end{tcolorbox}

This prompt enforces strict grounding: any query that exceeds basic chemistry knowledge triggers a retrieval tool call, and empty or irrelevant retrieval results produce an explicit "not found" response rather than a hallucinated answer.

Two generation backends are supported and are interchangeable via configuration:

\begin{itemize}
\item \textbf{Ollama}: Local inference server supporting open-weight models.
\item \textbf{OpenRouter}: Cloud-based API gateway providing access to a broad model catalog with configurable selection by price or throughput.
\end{itemize}

In this study, we report experiments using three locally hosted LLMs served via Ollama: Mistral-Small-3.2-24B, GPT-OSS-20B, and Qwen-3-8B. The models were used with their default inference parameters (see Table~\ref{tab:llm_config}).

\begin{table}[t]
\centering
\caption{Inference configurations of the locally hosted LLMs.}
\label{tab:llm_config}
\begin{tabular}{lccc}
\toprule
\textbf{Parameter} & \textbf{Mistral-Small-3.2-24B} & \textbf{GPT-OSS-20B} & \textbf{Qwen-3-8B} \\
\midrule
Temperature       & 0.15      & 1.0       & 0.6       \\
Top-$p$           & 0.9       & 0.9       & 0.95      \\
Top-$k$           & 40        & 40        & 20        \\
Min-$p$           & 0.0       & 0.0       & 0.0       \\
Repeat penalty    & 1.1       & 1.1       & 1.0       \\
Max tokens        & Unlimited & Unlimited & Unlimited \\
Quantization      & Q4\_K\_M  & Q4\_K\_M  & Q4\_K\_M  \\
Thinking mode     & Disabled  & Enabled   & Enabled   \\
\bottomrule
\end{tabular}
\end{table}

Responses are streamed token-by-token to the frontend via Server-Sent Events (SSE), minimising perceived latency. A lightweight Next.js/React web interface serves as the user-facing client.

\subsection{Configuration and Deployment}

The system is fully containerized and deployed via Docker Compose. All infrastructure services run as isolated containers with well-defined inter-service communication, enabling reproducible deployment on a single host or a small cluster. Table \ref{tab:deployment} lists each deployed service, its technology, default exposed port, and function within the system.

\begin{table}[h]
\centering
\caption{Deployment architecture of the RAG system, summarizing core services, underlying technologies, and their functional roles.}
\label{tab:deployment}
\begin{adjustbox}{width=\textwidth}
\begin{tabular}{p{2cm} p{2cm} p{8cm}}
\toprule
\textbf{Service} & \textbf{Technology} & \textbf{Function} \\
\midrule
Application server & Python (FastAPI / asyncio) & Hosts the RAG pipeline, REST API, and SSE streaming endpoint \\
Vector database & PostgreSQL 15 + pgvector & Stores document chunks, embeddings (text 1024-dim, image 512-dim), and article metadata; serves HNSW KNN queries \\
Full-text search engine & Meilisearch & Maintains an inverted index over all text chunks for sparse retrieval \\
Object storage & MinIO (S3-compatible) & Persists original PDF files and extracted figure images; accessed via S3 API \\
PDF conversion server & Marker PDF & Converts uploaded PDF documents to structured Markdown and extracts embedded images \\
\bottomrule
\end{tabular}
\end{adjustbox}
\end{table}

The pipeline consists of two stages. During \textbf{ingestion}, PDFs are parsed into Markdown and images, embedded, and indexed across PostgreSQL (dense), Meilisearch (sparse), and MinIO (images). During \textbf{inference}, a user query triggers parallel dense and sparse retrieval, optional reranking, and response generation by the configured LLM backend.

All pipeline components — embedding models, retrieval depth, reranker, chunking parameters, and generation backend — are controlled via a YAML configuration profile without code modification, enabling systematic experimentation across configurations and document corpora.

\subsection{Hardware Requirements}

The system requires dedicated GPU hardware due to local LLM inference via Ollama. The minimum recommended configuration for reproducing the experiments is summarized below. Only one LLM is loaded at a time, so VRAM requirements are determined by the largest model in use. Embedding (JinaEmbV3) and reranking (Jina Reranker M0) are served via external APIs and do not contribute to local VRAM usage. Hardware requirements are summarized in Table \ref{tab:hardware}.

\begin{table}[h]
\centering
\caption{Minimum hardware requirements for reproducing experiments.}
\label{tab:hardware}
\begin{tabular}{p{0.14\linewidth} p{0.14\linewidth} p{0.65\linewidth}}
\toprule
\textbf{Resource} & \textbf{Minimum} & \textbf{Notes} \\
\midrule
GPU VRAM & 24 GB & GPT-OSS-20B and Mistral-Small-3.2-24B require $\sim$12--14 GB at 4-bit quantization \\
System RAM & 32 GB & $\sim$16--20 GB for Ollama model weights; remaining memory for PostgreSQL ($\sim$2 GB), Meilisearch ($\sim$1 GB), MinIO ($\sim$512 MB), Marker PDF server ($\sim$2 GB), and application server \\
CPU & 8 cores & Supports parallel ingestion, chunking, and serving of containerized services \\
Storage & 100 GB SSD & Raw PDFs, extracted images (MinIO), PostgreSQL data and HNSW indexes, and Meilisearch index \\
\bottomrule
\end{tabular}
\end{table}


\newpage
\section{\benchname{}}
\label{appendix:benchmark}

\subsection{Document Corpus}
\label{sec:corpus}
 
All questions in \benchname{} are grounded in facts from open-access scientific publications (Gold or Diamond access based on OpenAlex~\citep{priem2022openalex}), ensuring full reproducibility and enabling the inclusion of complete PDF documents in the retrieval corpus. The corpus spans 8 chemical domains selected to cover diverse data types (numerical measurements, molecular structures, spectroscopic data, reaction schemes) and research areas (Table \ref{tab:bench-domains}).

\begin{table}[h]
\caption{Chemical domains in \benchname{} with the number of source articles per domain. Each domain covers a distinct area of chemistry with characteristic data types.}
\label{tab:bench-domains}
\centering
\small
\begin{tabular}{p{3.4cm}p{6.7cm}cc}
\toprule
\textbf{Domain} & \textbf{Description} & \textbf{Articles} & \textbf{Questions} \\
\midrule
Nanoparticle Antibacterials &
Antibacterial activity of inorganic nanoparticles against standard and multidrug-resistant bacterial strains, characterized by MIC and zone-of-inhibition data. & 41 & 245 \\
\addlinespace
Antibiotic-NP Synergy &
Combined antibacterial effects of nanoparticles and conventional antibiotics, including synergy metrics (FIC, fold increase) across standard and MDR strains. & 40 & 254 \\
\addlinespace
Nanozymes &
Catalytic properties of enzyme-mimicking nanoparticles, covering composition, morphology, substrate specificity, and kinetic parameters. & 40 & 240 \\
\addlinespace
Benzimidazole Antibiotics &
Structures and minimum inhibitory concentrations of benzimidazole-based antibiotics against various bacterial species. & 40 & 239 \\
\addlinespace
Oxazolidinone Antibiotics &
Oxazolidinone-class antibiotic structures and inhibitory concentrations against various bacterial species, with extraction provenance metadata. & 40 & 246 \\
\addlinespace
Chelate Metal Complexes &
Thermodynamic stability constants (log\,K) of metal-ligand chelate complexes (Ga, Gd, Tc, Lu) for MRI contrast agent design. & 62 & 275 \\
\addlinespace
Eye Drops &
Ophthalmic therapy of the corneal surface and tissues. & 30 & 263 \\
\addlinespace
Supramolecular Drug Systems &
Photostability of pharmaceutical co-crystals and cyclodextrin inclusion complexes, relating crystal form and host-guest chemistry to degradation resistance. & 40 & 245 \\
\bottomrule
\end{tabular}
\end{table}

Although all domains fall within biomedical and materials chemistry, they differ substantially in RAG-relevant characteristics such as document length, corpus size, and answer structure.

The retrieval database contains 333 articles: 224 directly referenced by at least one question, and 109 additional topically related articles that serve as retrieval distractors.

\subsection{Question Scenarios}
\label{sec:scenarios}
 
Questions are organized into six scenarios targeting distinct reasoning capabilities, from simple fact extraction to multi-turn dialogue.
 
\begin{enumerate}
    \item \textbf{Direct Fact Retrieval (Single-hop).} Extraction of a specific value or statement from a single source. Tests basic retrieval and information extraction. Requires a standard RAG configuration.
 
    \item \textbf{Sequential Retrieval (Multi-hop).} The answer to a first sub-question serves as the search key for a second source. Tests query decomposition and chained reasoning.
 
    \item \textbf{Conditional Retrieval.} Multiple conditions from different sources must be satisfied simultaneously. Tests cross-document reasoning and set intersection.
 
    \item \textbf{Comparison.} Entities extracted from separate sources must be compared quantitatively or qualitatively. Tests parallel retrieval and comparative reasoning.
 
    \item \textbf{Aggregation (Set Query).} All entities satisfying a condition must be collected from across the corpus. Tests retrieval completeness and aggregation.
 
    \item \textbf{Multi-turn Dialogue.} A sequence of thematically linked questions that may draw on different source articles within the same domain. Consecutive questions may share a topic, shift to a related subtopic, or address the same entity from a different angle. Tests context-dependent retrieval and coherent reasoning across topic transitions.
\end{enumerate}
 
 
\subsection{Question Modifications}
\label{sec:modifications}
 
Orthogonally to scenarios, each question may carry one or more of three modification types. These modifications introduce additional complexity without changing the underlying reasoning pattern.
 
\paragraph{Multimodal.} The answer is obtainable only from a non-textual source and is not duplicated in the article body. We distinguish two subtypes: \emph{table}, where the answer must be read from a structured data table, and \emph{image}, which covers all visual evidence (spectra, microscopy images, reaction schemes, plots).

\paragraph{Constrained.} The question includes metadata-level conditions that restrict which sources are relevant: \emph{theme} constraints specify a topical focus (e.g., ``in studies on silver nanoparticles for wound healing''), \emph{temporal} constraints reference publication dates or time periods of reported experiments, and \emph{source} constraints target specific journals, authors, or article titles.

\paragraph{Negative (Null/Quality).} Questions designed to probe robustness: \emph{conflicting sources} present contradictory answers across papers; \emph{incorrect queries} contain factually invalid premises; \emph{unanswerable queries} have no answer in the corpus. This tests hallucination resistance and appropriate abstention.

The full distribution of questions across the scenario-modification matrix is reported in Table~\ref{tab:matrix}.

\begin{table}[h]
\caption{Distribution of questions across the scenario-modification matrix (\% of total within each scenario). ``Base'' denotes questions without additional modifications. Total count shown as absolute number.}
\label{tab:matrix}
\centering
\small
\begin{tabular}{lcccc|c}
\toprule
\textbf{Scenario} & \textbf{Base (\%)} & \textbf{Multimodal (\%)} & \textbf{Constrained (\%)} & \textbf{Negative (\%)} & \textbf{Total} \\
\midrule
1. Single-hop       & 47.5 & 29.9 & 25.1 & 8.4 & 646 \\
2. Multi-hop        & 20.3 & 42.4 & 53.8 & 9.5 & 158 \\
3. Conditional      & 39.1 & 33.7 & 39.1 & 9.5 & 169 \\
4. Comparison       & 35.5 & 45.8 & 21.7 & 13.3 & 166 \\
5. Aggregation      & 49.1 & 17.4 & 41.9 & 4.2 & 167 \\
6. Multi-turn       & 42.2 & 29.5 & 31.8 & 6.4 & 701 \\
\midrule
\textbf{Total}      & 42.0 & 31.3 & 32.0 & 7.9 & 2007 \\
\bottomrule
\end{tabular}
\end{table}

\newpage
\section{Corpus-Level Features}
\label{appendix:dataset-features}

We describe each document collection using a fixed set of deterministic corpus-level features. These features are computed once per domain from the processed Markdown articles and the corresponding chunk embeddings. They are designed to capture corpus scale, document organization, linguistic complexity, and semantic structure, and are used as dataset descriptors for the surrogate model.

For each dataset, let $N$ denote the number of documents, $L_i$ the length of document $i$ in tokens, $C_i$ its length in characters, and $\mathbf{e}_i$ the normalized document embedding obtained by averaging all valid chunk embeddings belonging to that document.

\subsection{Statistical Features}

Document and size features quantify the amount of textual evidence available to the RAG system and the variability of document lengths across the corpus.

\paragraph{Number of Documents.}
This feature is defined as the total number of article records in the dataset. It measures the breadth of the corpus and determines how many source documents may be retrieved for a question.

\paragraph{Token-Based Document Length.}
We compute the mean and standard deviation of document lengths measured in tokens:
\begin{equation}
\bar{L} = \frac{1}{N}\sum_{i=1}^{N} L_i.
\end{equation}
Together, these statistics capture the typical amount of text per article and the degree of length heterogeneity in the corpus.

\paragraph{Total Token Corpus Size.}
The total token corpus size is the sum of token counts across all documents, $\sum_i L_i$. Larger corpora increase the retrieval search space and may require different retrieval depths or chunking strategies.

\paragraph{Character-Based Corpus Size.}
We also compute the mean, standard deviation, and total corpus size at the character level. These statistics are included because character length can reflect formatting density, formula-heavy text, and extraction artifacts that are not fully captured by token counts.

\subsection{Structure Features}

Structure features characterize how scientific information is organized inside documents. They reflect the presence of tables, figures, sections, and references, all of which can influence retrieval behavior and answer grounding.

\paragraph{Tables.}
We measure both the fraction of documents containing at least one detected table and the average number of tables per document. Tables are detected from Markdown and HTML table patterns. These features indicate how often information is presented in structured tabular form rather than plain prose.

\paragraph{Figures.}
We measure both the fraction of documents containing at least one detected figure and the average number of figures per document. Figures are counted using figure captions and non-artifact image links. These features approximate the degree of multimodal content in the corpus.

\paragraph{Sections.}
We compute the average number of meaningful Markdown headings per document after excluding metadata headings such as author information, acknowledgements, funding statements, and supplementary material. This feature captures the internal segmentation of scientific articles.

\paragraph{References.}
We compute the average number of bibliography entries detected in the references section of each document. This provides a proxy for article type and scientific context density.

\subsection{Linguistic Features}

Linguistic features describe textual complexity and domain specificity. They are computed from normalized document text using sentence tokenization, lemmatization, and domain-specific named entity recognition.

\paragraph{Sentence Length.}
We compute the mean number of tokens per sentence, averaged across documents. Longer sentences often indicate denser scientific writing and may make both retrieval matching and generation more challenging.

\paragraph{Vocabulary Size.}
Vocabulary size is defined as the number of unique lemmas observed across the dataset. It measures lexical diversity and helps distinguish narrow, repetitive corpora from heterogeneous corpora containing many distinct terms.

\paragraph{Technical Term Density.}
We estimate the average fraction of alphabetic word tokens that belong to recognized scientific or technical term spans. Formula-like spans are excluded so that textual terminology and symbolic notation are not conflated.

\paragraph{Chemical Name Density.}
We compute the average fraction of alphabetic word tokens that are identified as chemical entities by the chemical named entity recognizer. This feature captures how strongly the corpus is centered on explicit chemical compounds and materials.

\paragraph{Chemical Formula Density.}
We compute the fraction of lexical tokens that match formula-like patterns, such as molecular formulas or compact chemical identifiers. This captures symbolic chemical notation that may be important for exact retrieval.

\paragraph{Formula Density.}
We measure the fraction of document characters covered by detected LaTeX expressions. This reflects the amount of mathematical or symbolic notation in the extracted text.

\subsection{Embedding Features}

Embedding features summarize the semantic geometry of the dataset. Chunk embeddings are first parsed from the indexed chunk records. For each article, all valid chunk embeddings are averaged to obtain a document-level vector, and the resulting document vectors are normalized before computing corpus-level statistics.

\paragraph{Embedding Coverage.}
We count how many documents have at least one valid chunk embedding. This feature verifies embedding coverage and distinguishes missing embeddings from genuinely small corpora.

\paragraph{Embedding Variance.}
We compute the mean variance of document embeddings across embedding dimensions. Higher values indicate greater semantic spread among documents, while lower values suggest a more homogeneous corpus.

\paragraph{Average Pairwise Similarity.}
We compute the mean cosine similarity between all pairs of document embeddings:
\begin{equation}
\frac{2}{M(M-1)} \sum_{i<j} \cos(\mathbf{e}_i, \mathbf{e}_j),
\end{equation}
where $M$ is the number of documents with valid embeddings. This feature measures semantic coherence of the corpus.

\paragraph{Semantic Cluster Count.}
We estimate the number of semantic clusters in the dataset using $k$-means. Candidate values of $k$ are evaluated up to a maximum of 8, and the final value is selected using an elbow criterion based on the inertia curve.

\paragraph{Cluster Separation.}
For the selected clustering, we report the silhouette score when at least two clusters are present. This statistic measures how well separated the inferred semantic groups are.

\newpage
\section{Evaluation Metrics}
\label{appendix:metrics}

We evaluate the performance of the selected RAG system configurations using a suite of deterministic metrics that assess both retrieval quality and answer generation. These metrics are computationally efficient, reproducible, and do not rely on external language model judgments, ensuring stability across repeated runs.

The evaluation protocol is structured along the two main stages of the RAG pipeline: \textit{retrieval} and \textit{generation}.

\subsection{Retrieval Metrics}

Retrieval metrics quantify the ability of the system to identify and rank relevant scientific documents and text segments. Poor retrieval performance directly propagates to downstream hallucinations in generation.

\paragraph{Document-Level Retrieval}

We first evaluate whether the retriever successfully identifies relevant source documents:

\begin{itemize}
    \item \textbf{Golden DOI Recall.} Measures the fraction of ground-truth documents (identified by DOI) that are retrieved. Formally, this is the recall between the set of gold DOIs and the DOIs associated with retrieved chunks. Values range from 0 to 1.
    \item \textbf{Golden DOI MRR (Mean Reciprocal Rank).} Evaluates how highly the first relevant document is ranked:
\begin{equation}
\text{MRR} = \frac{1}{\text{rank}},
\end{equation}
where \textit{rank} denotes the position of the first correct document. This metric is particularly informative for reranking quality.
\end{itemize}







\subsection{Deterministic Generation Metrics}

Generation metrics evaluate the quality of model outputs with respect to a reference answer. Unlike traditional metrics that penalize verbosity, our evaluation emphasizes \textbf{recall-oriented completeness}, reflecting the importance of capturing all relevant scientific facts.

\paragraph{Fact-Level Completeness}
\begin{itemize}
    \item \textbf{Scientific Fact Recall.} The primary metric for generation quality. It decomposes the reference answer into atomic statements and uses a Natural Language Inference (NLI) model to estimate whether each statement is entailed by the generated output. This approach ensures robustness to paraphrasing while penalizing factual inconsistencies.
\end{itemize}


\paragraph{Semantic and Lexical Completeness}
\begin{itemize}
    \item \textbf{BERTScore Recall.} Computes semantic similarity between reference and generated text using contextual embeddings. Unlike strict fact-based metrics, it tolerates paraphrasing but may overlook subtle factual discrepancies.
    \item \textbf{ROUGE-L Recall.} Captures the longest common subsequence between reference and generated text, reflecting structural alignment and ordering of information.
    \item \textbf{Cosine Similarity.} Measures the cosine similarity between embedding representations of the generated and reference answers. This metric provides a coarse-grained estimate of semantic alignment, capturing overall similarity in meaning but being less sensitive to fine-grained factual correctness.
\end{itemize}

\newpage
\section{ML Pipeline}
\label{appendix:ml}

\paragraph{Data Processing}
Raw experimental tables containing RAG configuration parameters and evaluation scores are split into input features and a multi-dimensional target matrix. For each dataset, the data are randomly shuffled and split into training and test subsets using an 80/20 split with a fixed random seed of 42.

\paragraph{Model Training}
The task is formulated as multi-output regression. The pipeline trains 5 ensemble-based candidate models: CatBoost, XGBoost, LightGBM, ExtraTrees, and Random Forest. A shared preprocessing stage is applied before model fitting: numerical features are standardized with StandardScaler, while categorical features are encoded using one-hot encoding. Models that do not natively support multi-output regression are wrapped with MultiOutputRegressor, so that one regressor is trained per target metric. Each candidate model is evaluated using three-fold shuffled cross-validation on the training set. The optimization criterion is RMSE averaged over all predicted targets.

\paragraph{Best Model Selection}

Model selection is performed in two stages. First, all candidate model families are ranked by cross-validated RMSE. The top-performing models are then passed to the hyperparameter tuning stage using grid search (parameter grids are represented in Table \ref{tab:parameter_grids}). Hyperparameter tuning uses the same three-fold cross-validation protocol and minimizes RMSE. After tuning, the model with the lowest cross-validated RMSE is selected as the final model. To estimate the stability of the selected configuration, the best tuned model is refitted 5 times with different random seeds, and the pipeline reports mean and standard deviation of train and test RMSE, MAE, and R².

\begin{table}[h]
\centering
\caption{Hyperparameter grids used for model tuning.}
\label{tab:parameter_grids}
\begin{tabular}{lll}
\toprule
\textbf{Model} & \textbf{Parameter} & \textbf{Values} \\
\midrule

CatBoost
& \texttt{iterations} & 250, 400, 600 \\
& \texttt{learning\_rate} & 0.03, 0.06 \\
& \texttt{depth} & 3, 4, 6, 8 \\
& \texttt{l2\_leaf\_reg} & 1, 3 \\

\midrule

XGBoost
& \texttt{n\_estimators} & 250, 400, 600 \\
& \texttt{learning\_rate} & 0.03, 0.06 \\
& \texttt{max\_depth} & 2, 3, 4, 6 \\
& \texttt{subsample} & 0.85, 1.0 \\

\midrule

LightGBM
& \texttt{n\_estimators} & 250, 400, 600 \\
& \texttt{learning\_rate} & 0.03, 0.06 \\
& \texttt{num\_leaves} & 7, 15, 31, 63 \\
& \texttt{min\_child\_samples} & 10, 20 \\

\midrule

ExtraTrees
& \texttt{n\_estimators} & 300, 500, 800 \\
& \texttt{max\_depth} & None, 8, 16, 24 \\
& \texttt{min\_samples\_leaf} & 1, 2 \\
& \texttt{max\_features} & sqrt, 1.0 \\

\midrule

RandomForest
& \texttt{n\_estimators} & 300, 500, 800 \\
& \texttt{max\_depth} & None, 8, 16, 24 \\
& \texttt{min\_samples\_leaf} & 1, 2 \\
& \texttt{max\_features} & sqrt, 1.0 \\

\bottomrule
\end{tabular}
\end{table}

\newpage
\section{Surrogate Modeling Results}
\label{appendix:surrogate-results}

\paragraph{In-domain.} Following the pipeline described in Appendix \ref{appendix:ml}, for each of the three domains, we train ML model on RAG parameter configurations to predict RAG evaluation metrics on the same dataset it was trained. The results are summarized in Table \ref{tab:surrogate-in-domain}. The models achieve strong predictive performance reaching $R^2 \geq 0.84$ across all domains and metrics.

\begin{table}[h]
\centering
\caption{Results of in-domain RAG configuration space modeling within OA, MC and NZ domains reported as $R^2$ values. All standard deviations are $<0.01$.}
\label{tab:surrogate-in-domain}
\resizebox{\textwidth}{!}{%
\begin{tabular}{@{}ll
>{\centering\arraybackslash}p{2.2cm}
>{\centering\arraybackslash}p{2.2cm}
>{\centering\arraybackslash}p{1.8cm}
>{\centering\arraybackslash}p{1.8cm}
>{\centering\arraybackslash}p{2.2cm}
>{\centering\arraybackslash}p{2.5cm}@{}}
\toprule
\textbf{Train dataset} & \textbf{Best model} & \textbf{BERTScore Recall} & \textbf{Cosine Similarity} & \textbf{Golden DOI MRR} & \textbf{Golden DOI Recall} & \textbf{ROUGE-L Recall} & \textbf{Scientific Fact Recall} \\ \midrule
MC & LightGBM & $0.89$ & $0.87$ & $0.92$ & $0.97$ & $0.86$ & $0.94$ \\
NZ & XGBoost & $0.97$ & $0.95$ & $0.97$ & $0.99$ & $0.96$ & $0.97$ \\
OA & CatBoost & $0.84$ & $0.84$ & $0.91$ & $0.96$ & $0.85$ & $0.95$ \\ 
\bottomrule
\end{tabular}%
}
\end{table}

Since a random train/test split may place highly similar configurations in both sets, resulting in optimistic estimates of generalization due to interpolation between neighboring configurations, we additionally evaluated the models using leave-one-parameter-out (LOPO) experiments for the numerical parameters (\texttt{chunk\_size}, \texttt{dense\_k}, and \texttt{sparse\_k}). In each experiment, all configurations containing a specific parameter value were held out from training and used exclusively for testing. We did not apply LOPO to \texttt{search\_mode}, \texttt{reranker}, or \texttt{llm\_model}, as this would require predicting unseen categorical values, which is not a meaningful evaluation for tree-based surrogate models. As shown in Table~\ref{tab:lopo}, the surrogate generalizes well to unseen values of \texttt{chunk\_size} and \texttt{sparse\_k}, achieving performance comparable to that of the 80/20 random split reported in Table~\ref{tab:surrogate-in-domain}. The results for \texttt{dense\_k} are more mixed, particularly for retrieval metrics. This is consistent with our SHAP-IQ analysis, which identified \texttt{dense\_k} as one of the most influential parameters. Nevertheless, the surrogate maintains strong predictive performance  even under this more challenging evaluation, suggesting that it captures meaningful structure in the RAG design space rather than merely interpolating between observed configurations.

\begin{table}[h]
\centering
\caption{Performance of in-domain surrogate models under leave-one-parameter-out (LOPO) evaluation within OA, MC and NZ domains reported as $R^2$ values. In each experiment, all configurations containing a particular parameter value were excluded from training and used exclusively for testing. All standard deviations are $<0.01$.}
\label{tab:lopo}
\resizebox{\textwidth}{!}{%
\begin{tabular}{@{}l
>{\centering\arraybackslash}p{2.2cm}
>{\centering\arraybackslash}p{2.2cm}
>{\centering\arraybackslash}p{2.2cm}
>{\centering\arraybackslash}p{1.8cm}
>{\centering\arraybackslash}p{1.8cm}
>{\centering\arraybackslash}p{2.2cm}
>{\centering\arraybackslash}p{2.5cm}@{}}
\toprule
    \textbf{Dataset} & \textbf{Held-out parameter} & \textbf{BERT Score Recall} & \textbf{Cosine Similarity} & \textbf{Golden DOI MRR} & \textbf{Golden DOI Recall} & \textbf{ROUGE-L Recall} & \textbf{Scientific Fact Recall} \\ \midrule
& \texttt{dense\_k}    & 0.37              & 0.32              & 0.70           & 0.34              & 0.26           & 0.84                   \\
OA & \texttt{sparse\_k}   & 0.78              & 0.73              & 0.77           & 0.66              & 0.68           & 0.91                   \\
& \texttt{chunk\_size} & 0.76              & 0.73              & 0.87           & 0.93              & 0.74           & 0.94                   \\ \midrule
& \texttt{dense\_k}    & 0.54              & 0.34              & -2.30          & -2.30             & 0.65           & 0.87                   \\
MC & \texttt{sparse\_k}   & 0.87              & 0.82              & 0.87           & 0.86              & 0.85           & 0.94                   \\
& \texttt{chunk\_size} & 0.80              & 0.74              & 0.82           & 0.94              & 0.74           & 0.92                   \\ \midrule
& \texttt{dense\_k}    & 0.74              & 0.26              & -7.02          & -3.52             & 0.23           & 0.71                   \\
NZ & \texttt{sparse\_k}   & 0.95              & 0.92              & 0.88           & 0.85              & 0.90           & 0.95                   \\
& \texttt{chunk\_size} & 0.93              & 0.88              & 0.95           & 0.97              & 0.90           & 0.94                   \\ \bottomrule
\end{tabular}%
}
\end{table}

\begin{figure}[h!]
    \centering
    \includegraphics[width=\textwidth]{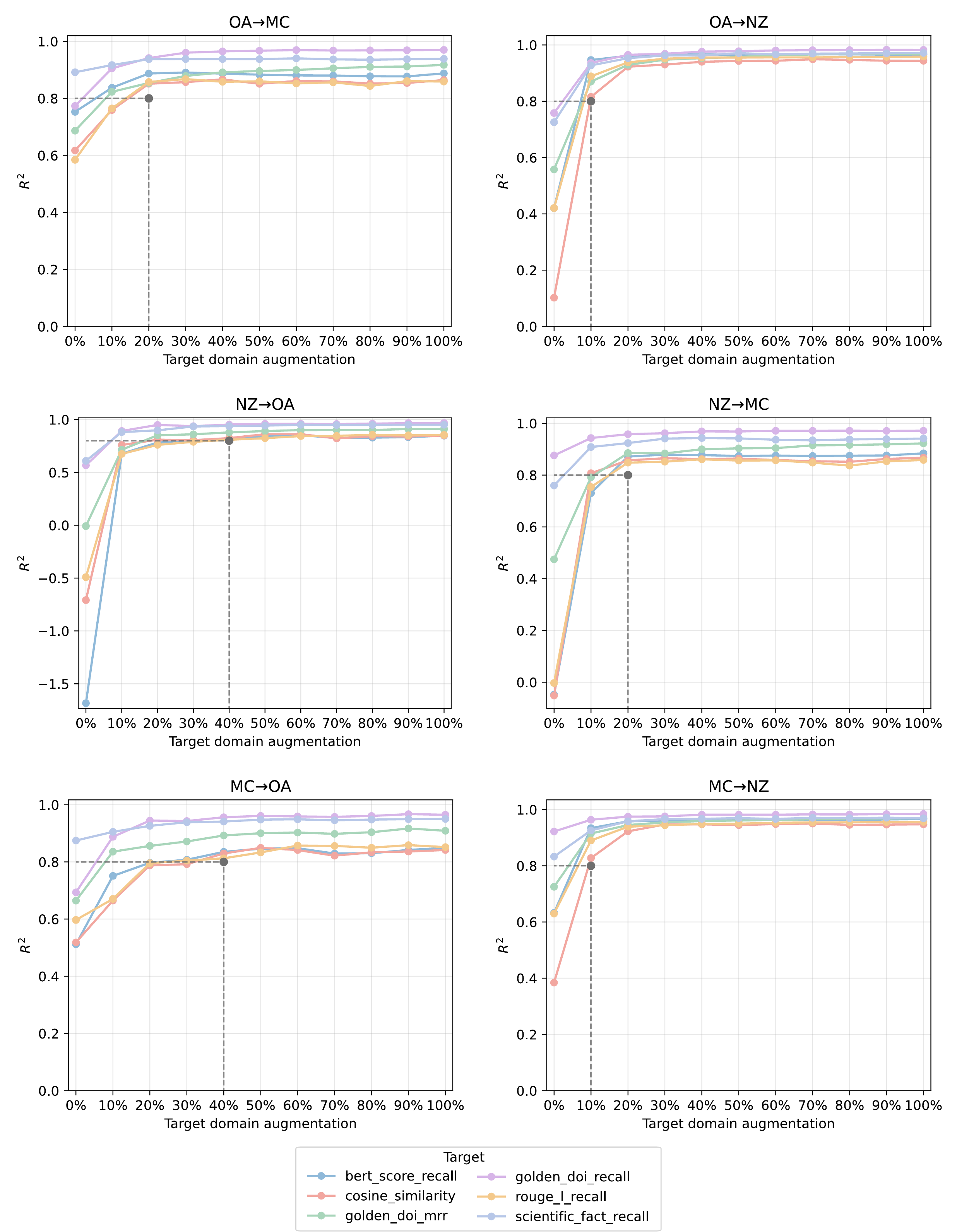}
    \caption{Augmented cross-domain transfer performance across all domain pairs. Each subplot shows $R^2$ as a function of the fraction of target-domain data added to a full source-domain training set, evaluated separately for each RAG evaluation metric. The dashed vertical and horizontal lines intersect at the minimum fraction of target-domain data required to achieve $R^2 > 0.8$ across all metrics.}
    \label{fig:augmented-transfer}
\end{figure}

\paragraph{Augmented cross-domain transfer.} To enable the surrogate to generalize across domains, we augment the RAG configuration features with PCA-compressed corpus-level descriptors that provide the model with a compact representation of domain characteristics. We train a LightGBM model on one full source-domain dataset and incrementally add an increasing fraction of target-domain experiments, evaluating predictive performance at each step. A consistent model architecture with default hyperparameters is used across all runs to ensure comparability. The results for all domain pairs are shown in Figure~\ref{fig:augmented-transfer}, with the minimum fraction of target-domain data required to achieve $R^2 > 0.8$ across all metrics highlighted for each pair.

\newpage
\section{Impact of Configuration Space Changes on Surrogate Model Interpretability}
\label{appendix:shapiq-stability}

To examine the robustness of surrogate-based interpretations to changes in the RAG configuration space, we extend the original experimental design by introducing Qwen-3-8B as an additional option for the generation model. All other configuration parameters and their possible values remain unchanged.

\begin{figure}[h]
    \centering
    \includegraphics[width=\textwidth]{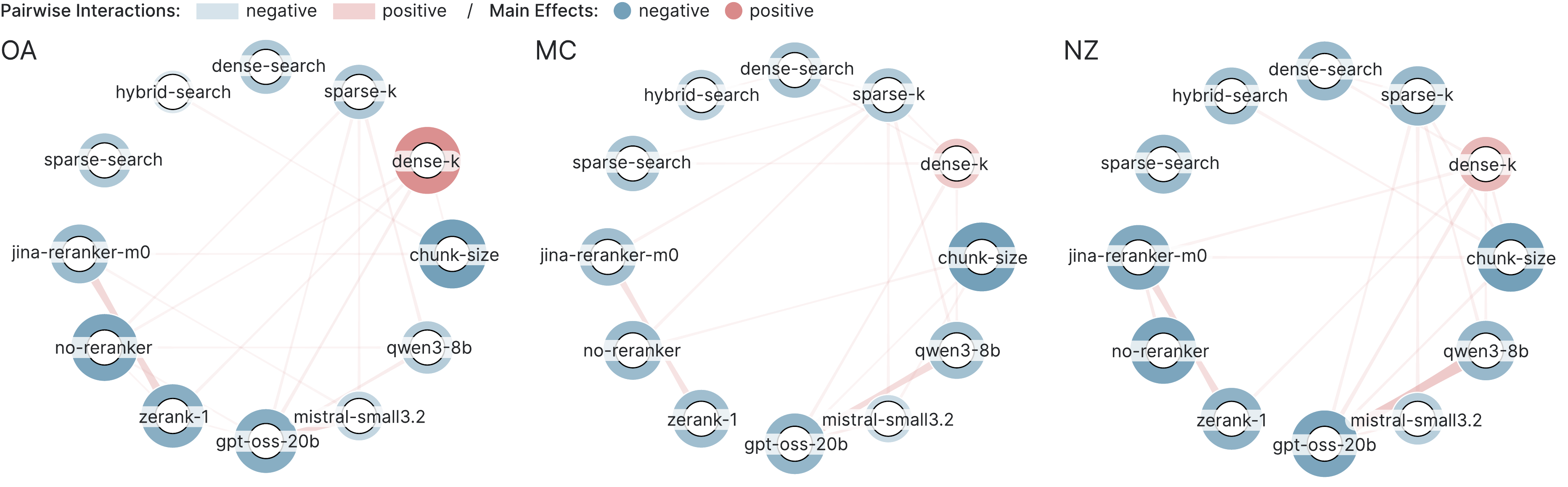}
    \caption{SHAP-IQ feature interaction networks for the extended RAG configuration space across three chemistry domains (OA, MC, and NZ). The extended space includes Qwen-3-8B as an additional generation model, while all other configuration parameters remain unchanged. }
    \label{fig:shap-qwen}
\end{figure}

We first train surrogate models on the extended configuration space across all three chemistry domains (OA, MC, and NZ), following the same training and evaluation protocol as in the original experiments. The resulting predictive performance is reported in Table~\ref{tab:surrogate-in-domain-extended}. We then apply SHAP-IQ to the trained surrogates to quantify individual feature contributions and pairwise interactions, with the corresponding interaction networks presented in Figure~\ref{fig:shap-qwen}.

\begin{table}[h]
\centering
\caption{Results of in-domain RAG configuration space modeling within OA, MC and NZ domains \textbf{under extended configuration space} reported as $R^2$ values. All standard deviations are $<0.01$.}
\label{tab:surrogate-in-domain-extended}
\resizebox{\textwidth}{!}{%
\begin{tabular}{@{}ll
>{\centering\arraybackslash}p{2.2cm}
>{\centering\arraybackslash}p{2.2cm}
>{\centering\arraybackslash}p{1.8cm}
>{\centering\arraybackslash}p{1.8cm}
>{\centering\arraybackslash}p{2.2cm}
>{\centering\arraybackslash}p{2.5cm}@{}}
\toprule
\textbf{Train dataset} & \textbf{Best model} & \textbf{BERTScore Recall} & \textbf{Cosine Similarity} & \textbf{Golden DOI MRR} & \textbf{Golden DOI Recall} & \textbf{ROUGE-L Recall} & \textbf{Scientific Fact Recall} \\ \midrule
MC & LightGBM & $0.85$ & $0.77$ & $0.94$ & $0.97$ & $0.80$ & $0.91$ \\
NZ & XGBoost & $0.94$ & $0.80$ & $0.97$ & $0.98$ & $0.91$ & $0.95$ \\
OA & CatBoost & $0.79$ & $0.75$ & $0.92$ & $0.95$ & $0.80$ & $0.92$ \\ 
\bottomrule
\end{tabular}%
}
\end{table}

Comparison with the original SHAP-IQ networks (Figure~\ref{fig:shap}B) reveals changes in both feature contributions and pairwise interactions. In particular, \texttt{chunk-size} becomes a more prominent negative contributor, while positive interactions become less pronounced in OA and MC, whereas NZ exhibits a more connected interaction structure. Notably, these changes extend beyond the LLM-related features, despite all retrieval and reranking parameter options remaining unchanged. At the same time, several patterns persist: \texttt{dense-k} retains a positive main effect across all three domains, while LLM and reranker configurations continue to exhibit predominantly negative main effects, highlighting again that their contributions to RAG performance are not uniformly beneficial. These findings suggest that the direction of certain feature contributions is relatively robust, whereas their magnitudes and interactions are more sensitive to the composition of the configuration space. Thus, surrogate-based analysis can uncover persistent dependencies among RAG parameters, although quantitative importance estimates should be interpreted in the context of the specific design space considered.

\newpage
\section{RAG Parameter Optimization}
\label{appendix:optimization}

\subsection{Implementation}

In our optimization experiments we explore the expanded RAG configuration space demonstrated in Table \ref{tab:expanded-space}. We treat the surrogates as objective functions within an Optuna-based multi-objective hyperparameter search, jointly optimizing for Golden DOI Recall and Scientific Fact Recall, running 2000 trials per experiment. From the obtained optimal configurations, we retain only those whose predicted performance exceeds that of the top configurations observed in the original RAG design space:

\begin{itemize}
    \item For \textbf{in-domain}, we consider two restriction levels: top-5 configurations observed in the original grid search for the given domain, and top-10 configurations that perform well across all three domains simultaneously. The former tests optimization within a domain-specific candidate pool, while the latter additionally enforces cross-domain robustness — identifying configurations that are strong regardless of domain.
    
    \item For \textbf{augmented transfer}, candidate configurations are restricted to the top-5 configurations from the original grid search for the target domain, with top-10 as a fallback when fewer than five candidates meet the predicted performance threshold. We restrict to target-domain candidates here — rather than cross-domain ones — since the surrogate is trained on multiple domains and is expected to have sufficient signal to identify strong target-specific configurations.
    
    \item For \textbf{out-of-domain transfer}, we apply the same restriction strategy as in the augmented setting: top-5 target-domain configurations, with top-10 as a fallback. This ensures consistency across transfer settings and allows for direct comparison of optimization quality across the two regimes.
\end{itemize}

To resolve ties between configurations with similar predicted scores (a common effect in tree-based models), we select the computationally cheaper option (e.g., configurations with fewer chunks passed to the LLM context) and additionally derive a robust configuration by taking the mode over parameter values across top-performing predictions.

\begin{table}[h]
\centering
\caption{Expanded hyperparameter search space explored in optimization experiments. 
The parameters \texttt{dense\_k} and \texttt{sparse\_k} are jointly active only in the \texttt{hybrid} retrieval mode.}
\label{tab:expanded-space}
\small
\begin{tabular}{ll}
\toprule
\textbf{Parameter} & \textbf{Search Space} \\
\midrule
\texttt{chunk\_size} &
$\{300, 400, 500,\dots,1600\}$ \\

\texttt{dense\_k} &
$\{5,6,7,\dots,30\}$ \\

\texttt{sparse\_k} &
$\{5,6,7,\dots,30\}$ \\

\texttt{search\_mode} &
$\{\texttt{dense-only}, \texttt{hybrid}, \texttt{sparse-only}\}$ \\

\texttt{reranker} &
$\{\texttt{jina-reranker-m0}, \texttt{none}, \texttt{zerank-1}\}$ \\

\texttt{llm\_model} &
$\{\texttt{gpt-oss-20b}, \texttt{mistral-small3.2}\}$ \\

\midrule
\textbf{Total configurations} & 61,152 \\
\bottomrule
\end{tabular}
\end{table}

\subsection{Results}

The obtained configurations for each datasets pair and optimization setup, as well as the deltas between the performance of an identified configuration evaluated on the real RAG system and a baseline from the original grid search are summarized in Tables \ref{tab:opt-1}, \ref{tab:opt-2}, \ref{tab:opt-3}, \ref{tab:opt-4}, \ref{tab:opt-5}, \ref{tab:opt-6}. Each row reports the delta between the performance of an identified configuration evaluated on the real RAG system and a baseline from the original grid search. Baselines and selection strategies differ for each run, as reflected in the 'Baseline' and 'Selection strategy' columns, respectively. Positive values indicate improvement over the baseline. The reported configurations were selected by maximizing the surrogate-predicted objective, and therefore small negative deltas after validation are expected due to surrogate prediction errors and the presence of several near-optimal configurations with very similar performance.

\begin{table}[h]
\centering
\caption{In-domain optimization results for OA domain, reported as deltas between the performance of an identified configuration and a baseline from the original grid search.}
\label{tab:opt-1}
\resizebox{\textwidth}{!}{%
\begin{tabular}{@{}
l
p{1cm}
>{\centering\arraybackslash}p{1.5cm}
p{4cm}
>{\centering\arraybackslash}p{1cm}
>{\centering\arraybackslash}p{1cm}
>{\centering\arraybackslash}p{1cm}
>{\centering\arraybackslash}p{1cm}
>{\centering\arraybackslash}p{1cm}
>{\centering\arraybackslash}p{1.1cm}
@{}}
\toprule
\textbf{Domain} & \textbf{Baseline} & \textbf{Selection strategy}  & \textbf{Configuration}                         & \textbf{Golden DOI Recall} & \textbf{Scientific Fact Recall} & \textbf{BERT Score Recall} & \textbf{Cosine Similarity} & \textbf{ROUGE -L Recall} & \textbf{Golden DOI MRR}\\ \midrule
OA   & top-10 all & cheapest & \begin{tabular}[c]{@{}l@{}}
search\_mode: hybrid \\
chunk\_size: 700 \\
dense\_k: 24 \\
sparse\_k: 27 \\
reranker: zerank-1 \\
llm\_model: mistral-small3.2 \\
chunk\_overlap: 175
\end{tabular}                                                                                                                         & +0,014              & +0,001 & +0,003 & +0,027 & +0,039 & +0,026                  \\ \midrule
OA   & top-10 all & mode     & \begin{tabular}[c]{@{}l@{}}
search\_mode: hybrid \\
chunk\_size: 900 \\
dense\_k: 19 \\
sparse\_k: 20 \\
reranker: zerank-1 \\
llm\_model: mistral-small3.2 \\
chunk\_overlap: 225
\end{tabular}                                                                                                                         & -0,029             & +0,002 & 0,000 & +0,029 & +0,012 & +0,016                 \\ \midrule
OA   & top-5 OA & cheapest   & \begin{tabular}[c]{@{}l@{}}
search\_mode: hybrid \\
chunk\_size: 500 \\
dense\_k: 19 \\
sparse\_k: 15 \\
reranker: zerank-1 \\
llm\_model: mistral-small3.2 \\
chunk\_overlap: 125
\end{tabular}                                                                                                                         & -0,017             & +0,001 & -0,046 & -0,068 & -0,095 & -0,045                  \\ \midrule
OA   & top-5 OA & mode       & \begin{tabular}[c]{@{}l@{}}
search\_mode: hybrid \\
chunk\_size: 600 \\
dense\_k: 26 \\
sparse\_k: 17 \\
reranker: zerank-1 \\
llm\_model: mistral-small3.2 \\
chunk\_overlap: 150
\end{tabular}                                                                                                                         & -0,007             & +0,002 & +0,023 & +0,044 & +0,066 & -0,014                \\ \bottomrule
\end{tabular}%
}
\end{table}

\begin{table}[h]
\centering
\caption{In-domain optimization results for MC domain, reported as deltas between the performance of an identified configuration and a baseline from the original grid search.}
\label{tab:opt-2}
\resizebox{\textwidth}{!}{%
\begin{tabular}{@{}
l
p{1cm}
>{\centering\arraybackslash}p{1.5cm}
p{4cm}
>{\centering\arraybackslash}p{1cm}
>{\centering\arraybackslash}p{1cm}
>{\centering\arraybackslash}p{1cm}
>{\centering\arraybackslash}p{1cm}
>{\centering\arraybackslash}p{1cm}
>{\centering\arraybackslash}p{1.1cm}
@{}}
\toprule
\textbf{Domain} & \textbf{Baseline} & \textbf{Selection strategy}  & \textbf{Configuration}                         & \textbf{Golden DOI Recall} & \textbf{Scientific Fact Recall} & \textbf{BERT Score Recall} & \textbf{Cosine Similarity} & \textbf{ROUGE -L Recall} & \textbf{Golden DOI MRR}\\ \midrule
MC & top-10 all & cheapest & \begin{tabular}[c]{@{}l@{}}
search\_mode: hybrid \\
chunk\_size: 700 \\
dense\_k: 20 \\
sparse\_k: 24 \\
reranker: zerank-1 \\
llm\_model: mistral-small3.2 \\
chunk\_overlap: 175
\end{tabular}
& -0,021 & +0,001 & +0,035 & +0,065 & +0,099 & +0,034 \\ \midrule
MC & top-10 all & mode & \begin{tabular}[c]{@{}l@{}}
search\_mode: hybrid \\
chunk\_size: 800 \\
dense\_k: 18 \\
sparse\_k: 20 \\
reranker: zerank-1 \\
llm\_model: mistral-small3.2 \\
chunk\_overlap: 200
\end{tabular}
& -0,021 & +0,001 & +0,025 & +0,045 & +0,087 & +0,037 \\ \midrule
MC & top-5 MC & cheapest & \begin{tabular}[c]{@{}l@{}}
search\_mode: hybrid \\
chunk\_size: 1100 \\
dense\_k: 17 \\
sparse\_k: 19 \\
reranker: zerank-1 \\
llm\_model: mistral-small3.2 \\
chunk\_overlap: 275
\end{tabular}
& -0,059 & 0,001 & +0,013 & +0,060 & +0,083 & +0,008 \\ \midrule
MC & top-5 MC & mode & \begin{tabular}[c]{@{}l@{}}
search\_mode: hybrid \\
chunk\_size: 1400 \\
dense\_k: 20 \\
sparse\_k: 20 \\
reranker: zerank-1 \\
llm\_model: mistral-small3.2 \\
chunk\_overlap: 350
\end{tabular}
& -0,028 & 0,000 & +0,018 & +0,062 & +0,068 & +0,007 \\ \bottomrule
\end{tabular}%
}
\end{table}

\begin{table}[h]
\centering
\caption{In-domain optimization results for NZ domain, reported as deltas between the performance of an identified configuration and a baseline from the original grid search.}
\label{tab:opt-3}
\resizebox{\textwidth}{!}{%
\begin{tabular}{@{}
l
p{1cm}
>{\centering\arraybackslash}p{1.5cm}
p{4cm}
>{\centering\arraybackslash}p{1cm}
>{\centering\arraybackslash}p{1cm}
>{\centering\arraybackslash}p{1cm}
>{\centering\arraybackslash}p{1cm}
>{\centering\arraybackslash}p{1cm}
>{\centering\arraybackslash}p{1.1cm}
@{}}
\toprule
\textbf{Domain} & \textbf{Baseline} & \textbf{Selection strategy}  & \textbf{Configuration}                         & \textbf{Golden DOI Recall} & \textbf{Scientific Fact Recall} & \textbf{BERT Score Recall} & \textbf{Cosine Similarity} & \textbf{ROUGE -L Recall} & \textbf{Golden DOI MRR}\\ \midrule
NZ & top-10 all & cheapest & \begin{tabular}[c]{@{}l@{}}
search\_mode: dense-only \\
chunk\_size: 800 \\
dense\_k: 20 \\
sparse\_k: 0 \\
reranker: zerank-1 \\
llm\_model: mistral-small3.2 \\
chunk\_overlap: 200
\end{tabular}
& -0,031 & -0,001 & -0,010 & -0,023 & +0,026 & -0,008 \\ \midrule
NZ & top-10 all & mode & \begin{tabular}[c]{@{}l@{}}
search\_mode: dense-only \\
chunk\_size: 800 \\
dense\_k: 27 \\
sparse\_k: 0 \\
reranker: zerank-1 \\
llm\_model: mistral-small3.2 \\
chunk\_overlap: 200
\end{tabular}
& +0,010 & -0,001 & -0,015 & -0,013 & +0,017 & +0,001 \\ \midrule
NZ & top-5 NZ & cheapest & \begin{tabular}[c]{@{}l@{}}
search\_mode: hybrid \\
chunk\_size: 800 \\
dense\_k: 20 \\
sparse\_k: 25 \\
reranker: zerank-1 \\
llm\_model: mistral-small3.2 \\
chunk\_overlap: 200
\end{tabular}
& 0,000 & -0,001 & +0,008 & -0,002 & +0,024 & -0,001 \\ \midrule
NZ & top-5 NZ & mode & \begin{tabular}[c]{@{}l@{}}
search\_mode: hybrid \\
chunk\_size: 800 \\
dense\_k: 25 \\
sparse\_k: 20 \\
reranker: zerank-1 \\
llm\_model: mistral-small3.2 \\
chunk\_overlap: 200
\end{tabular}
& +0,010 & -0,001 & -0,011 & -0,003 & -0,020 & -0,005 \\ \bottomrule
\end{tabular}%
}
\end{table}

\begin{table}[h]
\centering
\caption{In-domain optimization results for OA, MC, and NZ domains. Here, surrogate model trained on \textbf{all three domains} is used to optimize configurations for each domain. Each row reports the delta between the performance of an identified configuration and a baseline from the original grid search.}
\label{tab:opt-4}
\resizebox{\textwidth}{!}{%
\begin{tabular}{@{}
p{1.2cm}
p{1cm}
>{\centering\arraybackslash}p{1.5cm}
p{4cm}
>{\centering\arraybackslash}p{1cm}
>{\centering\arraybackslash}p{1cm}
>{\centering\arraybackslash}p{1cm}
>{\centering\arraybackslash}p{1cm}
>{\centering\arraybackslash}p{1cm}
>{\centering\arraybackslash}p{1.1cm}
@{}}
\toprule
\textbf{Target domain} & \textbf{Baseline} & \textbf{Selection strategy}  & \textbf{Configuration}                         & \textbf{Golden DOI Recall} & \textbf{Scientific Fact Recall} & \textbf{BERT Score Recall} & \textbf{Cosine Similarity} & \textbf{ROUGE -L Recall} & \textbf{Golden DOI MRR}\\ \midrule
OA & top-5 OA & cheapest & \begin{tabular}[c]{@{}l@{}}
search\_mode: hybrid \\
chunk\_size: 800 \\
dense\_k: 16 \\
sparse\_k: 25 \\
reranker: zerank-1 \\
llm\_model: mistral-small3.2 \\
chunk\_overlap: 200
\end{tabular}
& -0,024 & 0,000 & +0,004 & +0,029 & +0,036 & +0,008 \\ \midrule
OA & top-5 OA & mode & \begin{tabular}[c]{@{}l@{}}
search\_mode: hybrid \\
chunk\_size: 800 \\
dense\_k: 20 \\
sparse\_k: 25 \\
reranker: zerank-1 \\
llm\_model: mistral-small3.2 \\
chunk\_overlap: 200
\end{tabular}
& -0,024 & +0,002 & 0,000 & +0,038 & 0,001 & +0,020 \\ \midrule
MC & top-5 MC & cheapest & \begin{tabular}[c]{@{}l@{}}
search\_mode: hybrid \\
chunk\_size: 800 \\
dense\_k: 17 \\
sparse\_k: 20 \\
reranker: zerank-1 \\
llm\_model: mistral-small3.2 \\
chunk\_overlap: 200
\end{tabular}
& -0,031 & +0,001 & +0,029 & +0,049 & +0,093 & +0,053 \\ \midrule
MC & top-5 MC & mode & \begin{tabular}[c]{@{}l@{}}
search\_mode: hybrid \\
chunk\_size: 800 \\
dense\_k: 26 \\
sparse\_k: 20 \\
reranker: zerank-1 \\
llm\_model: mistral-small3.2 \\
chunk\_overlap: 200
\end{tabular}
& 0,000 & +0,001 & +0,025 & +0,075 & +0,088 & +0,056 \\ \midrule
NZ & top-5 NZ & cheapest & \begin{tabular}[c]{@{}l@{}}
search\_mode: hybrid \\
chunk\_size: 800 \\
dense\_k: 16 \\
sparse\_k: 20 \\
reranker: zerank-1 \\
llm\_model: mistral-small3.2 \\
chunk\_overlap: 200
\end{tabular}
& -0,010 & 0,000 & -0,005 & -0,007 & +0,006 & -0,006 \\ \midrule
NZ & top-5 NZ & mode & \begin{tabular}[c]{@{}l@{}}
search\_mode: hybrid \\
chunk\_size: 800 \\
dense\_k: 28 \\
sparse\_k: 16 \\
reranker: zerank-1 \\
llm\_model: mistral-small3.2 \\
chunk\_overlap: 200
\end{tabular}
& +0,010 & 0,000 & +0,001 & +0,003 & +0,040 & -0,015 \\ \bottomrule
\end{tabular}%
}
\end{table}

\begin{table}[h]
\centering
\caption{Augmented domain transfer optimization results for OA, MC, and NZ domains. Each row reports the delta between the performance of an identified configuration and a baseline from the original grid search.}
\label{tab:opt-5}
\resizebox{\textwidth}{!}{%
\begin{tabular}{@{}
p{1.3cm}
p{1.2cm}
p{1cm}
>{\centering\arraybackslash}p{1.5cm}
p{4cm}
>{\centering\arraybackslash}p{1cm}
>{\centering\arraybackslash}p{1cm}
>{\centering\arraybackslash}p{1cm}
>{\centering\arraybackslash}p{1cm}
>{\centering\arraybackslash}p{1cm}
>{\centering\arraybackslash}p{1.1cm}
@{}}
\toprule
\textbf{Train domains} & \textbf{Target domain} & \textbf{Baseline} & \textbf{Selection strategy}  & \textbf{Configuration}                         & \textbf{Golden DOI Recall} & \textbf{Scientific Fact Recall} & \textbf{BERT Score Recall} & \textbf{Cosine Similarity} & \textbf{ROUGE -L Recall} & \textbf{Golden DOI MRR}\\ \midrule
100\% OA+ 20\% MC & MC & top-5 MC & cheapest & \begin{tabular}[c]{@{}l@{}}
search\_mode: hybrid \\
chunk\_size: 700 \\
dense\_k: 24 \\
sparse\_k: 11 \\
reranker: zerank-1 \\
llm\_model: mistral-small3.2 \\
chunk\_overlap: 175
\end{tabular}
& -0,042 & +0,001 & +0,034 & +0,064 & +0,116 & +0,035 \\ \midrule
100\% OA+ 20\% MC & MC & top-5 MC & mode & \begin{tabular}[c]{@{}l@{}}
search\_mode: hybrid \\
chunk\_size: 800 \\
dense\_k: 16 \\
sparse\_k: 11 \\
reranker: zerank-1 \\
llm\_model: mistral-small3.2 \\
chunk\_overlap: 200
\end{tabular}
& -0,042 & +0,001 & +0,030 & +0,044 & +0,070 & +0,055 \\ \midrule
100\% NZ+ 20\% MC & MC & top-10 MC & cheapest = mode & \begin{tabular}[c]{@{}l@{}}
search\_mode: hybrid \\
chunk\_size: 1400 \\
dense\_k: 26 \\
sparse\_k: 20 \\
reranker: zerank-1 \\
llm\_model: mistral-small3.2 \\
chunk\_overlap: 350
\end{tabular}
& -0,028 & +0,002 & -0,066 & -0,094 & -0,136 & +0,027 \\ \midrule
100\% OA+ 10\% NZ & NZ & top-5 NZ & cheapest & \begin{tabular}[c]{@{}l@{}}
search\_mode: dense-only \\
chunk\_size: 1100 \\
dense\_k: 18 \\
sparse\_k: 0 \\
reranker: zerank-1 \\
llm\_model: mistral-small3.2 \\
chunk\_overlap: 275
\end{tabular}
& -0,038 & -0,001 & -0,020 & -0,011 & -0,039 & +0,010 \\ \midrule
100\% OA+ 10\% NZ & NZ & top-5 NZ & mode & \begin{tabular}[c]{@{}l@{}}
search\_mode: dense-only \\
chunk\_size: 1100 \\
dense\_k: 22 \\
sparse\_k: 0 \\
reranker: zerank-1 \\
llm\_model: mistral-small3.2 \\
chunk\_overlap: 275
\end{tabular}
& -0,038 & -0,001 & -0,005 & -0,005 & -0,019 & -0,004 \\ \midrule
100\% MC+ 10\% NZ & NZ & top-10 NZ & cheapest & \begin{tabular}[c]{@{}l@{}}
search\_mode: hybrid \\
chunk\_size: 1200 \\
dense\_k: 19 \\
sparse\_k: 29 \\
reranker: zerank-1 \\
llm\_model: mistral-small3.2 \\
chunk\_overlap: 300
\end{tabular}
& -0,007 & 0,000 & +0,001 & -0,013 & -0,041 & +0,013 \\ \midrule
100\% MC+ 10\% NZ & NZ & top-10 NZ & mode & \begin{tabular}[c]{@{}l@{}}
search\_mode: hybrid \\
chunk\_size: 1200 \\
dense\_k: 21 \\
sparse\_k: 29 \\
reranker: zerank-1 \\
llm\_model: mistral-small3.2 \\
chunk\_overlap: 300
\end{tabular}
& +0,003 & 0,000 & +0,013 & -0,039 & -0,067 & +0,021 \\ \bottomrule
\end{tabular}%
}
\end{table}

\begin{table}[h]
\centering
\caption{Out-of-domain transfer optimization results for OA, MC, and NZ domains. Each row reports the delta between the performance of an identified configuration and a baseline from the original grid search.}
\label{tab:opt-6}
\resizebox{\textwidth}{!}{%
\begin{tabular}{@{}
p{1.3cm}
p{1.2cm}
p{1cm}
>{\centering\arraybackslash}p{1.5cm}
p{4cm}
>{\centering\arraybackslash}p{1cm}
>{\centering\arraybackslash}p{1cm}
>{\centering\arraybackslash}p{1cm}
>{\centering\arraybackslash}p{1cm}
>{\centering\arraybackslash}p{1cm}
>{\centering\arraybackslash}p{1.1cm}
@{}}
\toprule
\textbf{Train domains} &\textbf{Target domain} & \textbf{Baseline} & \textbf{Selection strategy}  & \textbf{Configuration}                         & \textbf{Golden DOI Recall} & \textbf{Scientific Fact Recall} & \textbf{BERT Score Recall} & \textbf{Cosine Similarity} & \textbf{ROUGE -L Recall} & \textbf{Golden DOI MRR}\\ \midrule
100\% OA+ 100\% MC & NZ & top-10 all & cheapest & \begin{tabular}[c]{@{}l@{}}
search\_mode: hybrid \\
chunk\_size: 800 \\
dense\_k: 24 \\
sparse\_k: 25 \\
reranker: zerank-1 \\
llm\_model: mistral-small3.2 \\
chunk\_overlap: 200
\end{tabular}
& +0,021 & 0,000 & -0,005 & -0,010 & 0,003 & -0,014 \\ \midrule
100\% OA+ 100\% MC & NZ & top-10 all & mode & \begin{tabular}[c]{@{}l@{}}
search\_mode: hybrid \\
chunk\_size: 1000 \\
dense\_k: 24 \\
sparse\_k: 22 \\
reranker: zerank-1 \\
llm\_model: mistral-small3.2 \\
chunk\_overlap: 250
\end{tabular}
& 0,000 & 0,000 & +0,016 & 0,004 & +0,018 & -0,006 \\ \bottomrule
\end{tabular}%
}
\end{table}

\subsection{Comparison with Random Search}
\label{appendix:random-search}

To assess the effectiveness of in-domain surrogate-guided optimization relative to uninformed configuration selection, we introduce a random-search baseline in a simulated optimization setting. For each chemistry domain, we use the 648 RAG configurations with precomputed Golden DOI Recall and Scientific Fact Recall scores to simulate an optimization experiment under a limited evaluation budget. The configurations are partitioned into 148 initially known configurations and a pool of 500 candidates. The split is constructed such that exactly 10 of 500 candidates outperform the best initially known configuration in terms of the sum of the two recall metrics. During the simulation, the precomputed scores of the 500 candidates are treated as unknown and revealed only when a candidate is selected for evaluation.

Both methods are allocated an equal budget of 10 rel RAG evaluations per run, with 10 runs conducted using random seeds 0–9. Random search selects 10 uniformly sampled candidates. For surrogate-guided optimization, a domain-specific ML model is first cross-validated and fitted on the 148 initially known configurations, Optuna's TPE sampler then performs 1000 inexpensive surrogate-based trials, after which the 10 unique candidates with the highest predicted sums of the two recall metrics are selected for simulated evaluation. Their precomputed scores are retrieved to emulate real RAG evaluations. For both methods, the final configuration is selected based on the highest observed sum of the two metrics. A run is counted as a win if the selected configuration outperforms the best initially known configuration. We report the win rate across 10 runs, along with the mean and sample standard deviation of the two recall metrics for the final selected configurations.

As shown in Table~\ref{tab:random_search}, surrogate-guided optimization consistently outperforms random search across all three domains under the same real RAG evaluations budget. Random search achieves win rates of 30\%, 30\%, and 10\% for OA, MC, and NZ, respectively, whereas surrogate-guided optimization achieves 100\% across all domains. The surrogate-guided approach also yields higher average values for both target metrics, demonstrating its advantage over random configuration selection.

\begin{table}[t]
\centering
\caption{Comparison of surrogate-guided optimization
and random search across three chemistry domains under an equal 10 of ten real RAG evaluations per run. Results are reported as mean $\pm$ standard deviation.}
\label{tab:random_search}
\small
\begin{tabular}{llccc}
\toprule
\textbf{Domain} & \textbf{Method}
& \textbf{Win Rate (\%)}
& \textbf{Golden DOI Recall}
& \textbf{Scientific Fact Recall} \\
\midrule
\multirow{2}{*}{OA}
& Random Search & 30 & $0.86 \pm 0.07$ & $0.84 \pm 0.09$ \\
& Surrogate-guided & \textbf{100}
& $\mathbf{0.91 \pm 0.00}$ & $\mathbf{0.90 \pm 0.00}$ \\
\midrule
\multirow{2}{*}{MC}
& Random Search & 30 & $0.83 \pm 0.05$ & $0.83 \pm 0.15$ \\
& Surrogate-guided & \textbf{100}
& $\mathbf{0.89 \pm 0.00}$ & $\mathbf{0.94 \pm 0.00}$ \\
\midrule
\multirow{2}{*}{NZ}
& Random Search & 10 & $0.85 \pm 0.04$ & $0.89 \pm 0.12$ \\
& Surrogate-guided & \textbf{100}
& $\mathbf{0.90 \pm 0.00}$ & $\mathbf{0.98 \pm 0.00}$ \\
\bottomrule
\end{tabular}
\end{table}

\subsection{Human Study of Optimized Configurations}

To complement the automatic evaluation metrics, we qualitatively inspected some representative QA examples from the OA domain where surrogate-guided optimization changed the generated answers. As shown in Table~\ref{tab:qualitative_analysis}, the optimized configurations correctly recovered scientific facts that were missed by the baseline RAG system, including numerical values, confidence intervals, and compound identities. In some cases, neither configuration produced a fully correct answer, but the optimized system yielded responses closer to the reference answer.

\begin{table}[t]
\centering
\caption{Qualitative comparison of baseline and optimized RAG answers on questions from the OA domain.}
\label{tab:qualitative_analysis}
\small
\begin{tabular}{p{0.27\textwidth}p{0.18\textwidth}p{0.20\textwidth}p{0.24\textwidth}}
\toprule
\textbf{Question (short)} &
\textbf{Ground Truth} &
\textbf{Baseline RAG} &
\textbf{Optimized RAG} \\
\midrule
MIC of tedizolid against \textit{M. tuberculosis} &
0.25 $\mu$g/mL &
0.5 mg/L &
0.25 mg/L \\
\midrule
Tedizolid non-inferiority 95\% CI &
$-2.0$ to $6.5$ &
$-10\%$ to $10\%$ &
$-2.0$ to $6.5$ \\
\midrule
Oxazolidinone in Gao et al. (2023) crossover study &
Delpazolid &
Contezolid &
Delpazolid \\
\midrule
Antibiotic with $K_i = 182.26\,\mu$M and MIC = 2 $\mu$g/mL &
Linezolid &
Not identified &
Contezolid \\
\midrule
MIC values of vancomycin and linezolid &
Both 2 $\mu$g/mL &
Vancomycin $>256$ $\mu$g/mL &
Vancomycin 1 $\mu$g/mL; linezolid 2 $\mu$g/mL \\
\bottomrule
\end{tabular}
\end{table}


\clearpage
\newpage
\section*{NeurIPS Paper Checklist}

\begin{enumerate}

\item {\bf Claims}
    \item[] Question: Do the main claims made in the abstract and introduction accurately reflect the paper's contributions and scope?
    \item[] Answer: \answerYes{}
    \item[] Justification: Yes. All main claims made in the abstract and introduction are directly supported by experimental results. Surrogate model accuracy is reported in Table~\ref{tab:generalization}, cross-domain transfer results are shown in Figure~\ref{fig:augmented-transfer}, and optimization results are validated through real RAG experiments in Figure~\ref{fig:optimization}.

\item {\bf Limitations}
    \item[] Question: Does the paper discuss the limitations of the work performed by the authors?
    \item[] Answer: \answerYes{}
    \item[] Justification: Limitations are provided in section 5.4.

\item {\bf Theory assumptions and proofs}
    \item[] Question: For each theoretical result, does the paper provide the full set of assumptions and a complete (and correct) proof?
    \item[] Answer: \answerNA{}
    \item[] Justification: The paper does not include theoretical results.

    \item {\bf Experimental result reproducibility}
    \item[] Question: Does the paper fully disclose all the information needed to reproduce the main experimental results of the paper to the extent that it affects the main claims and/or conclusions of the paper (regardless of whether the code and data are provided or not)?
    \item[] Answer: \answerYes{}
    \item[] Justification: We provide all algorithms implementation details in the Appendix E and G, including parameter grids for models and random states.

\item {\bf Open access to data and code}
    \item[] Question: Does the paper provide open access to the data and code, with sufficient instructions to faithfully reproduce the main experimental results, as described in supplemental material?
    \item[] Answer: \answerYes{}
    \item[] Justification: We provide links to public repositories with datasets and code.

\item {\bf Experimental setting/details}
    \item[] Question: Does the paper specify all the training and test details (e.g., data splits, hyperparameters, how they were chosen, type of optimizer) necessary to understand the results?
    \item[] Answer: \answerYes{}
    \item[] Justification: We specify these details in Appendix E and G.

\item {\bf Experiment statistical significance}
    \item[] Question: Does the paper report error bars suitably and correctly defined or other appropriate information about the statistical significance of the experiments?
    \item[] Answer: \answerYes{}
    \item[] Justification: We report error bars for all the trained ML models.

\item {\bf Experiments compute resources}
    \item[] Question: For each experiment, does the paper provide sufficient information on the computer resources (type of compute workers, memory, time of execution) needed to reproduce the experiments?
    \item[] Answer: \answerYes{}
    \item[] Justification: We provide the information on the computer resources in Table 4.
    
\item {\bf Code of ethics}
    \item[] Question: Does the research conducted in the paper conform, in every respect, with the NeurIPS Code of Ethics \url{https://neurips.cc/public/EthicsGuidelines}?
    \item[] Answer: \answerYes{}
    \item[] Justification: The research conducted in the paper conforms, in every respect, with the NeurIPS Code of Ethics.

\item {\bf Broader impacts}
    \item[] Question: Does the paper discuss both potential positive societal impacts and negative societal impacts of the work performed?
    \item[] Answer: \answerNA{}
    \item[] Justification: This work presents a methodological framework for RAG system optimization and does not introduce new capabilities that could directly cause harm. The societal impact is expected to be limited to reduced computational cost of RAG system development, which has positive implications for research efficiency and accessibility.
    
\item {\bf Safeguards}
    \item[] Question: Does the paper describe safeguards that have been put in place for responsible release of data or models that have a high risk for misuse (e.g., pre-trained language models, image generators, or scraped datasets)?
    \item[] Answer: \answerNA{}
    \item[] Justification: The paper does not produce data or models that have a high risk for misuse.

\item {\bf Licenses for existing assets}
    \item[] Question: Are the creators or original owners of assets (e.g., code, data, models), used in the paper, properly credited and are the license and terms of use explicitly mentioned and properly respected?
    \item[] Answer: \answerYes{}
    \item[] Justification: The paper cites the authors of the used packages (Optuna, SHAP-IQ).

\item {\bf New assets}
    \item[] Question: Are new assets introduced in the paper well documented and is the documentation provided alongside the assets?
    \item[] Answer: \answerYes{}
    \item[] Justification: We provide comprehensive documentation for RAG framework and experimental code in our repositories.

\item {\bf Crowdsourcing and research with human subjects}
    \item[] Question: For crowdsourcing experiments and research with human subjects, does the paper include the full text of instructions given to participants and screenshots, if applicable, as well as details about compensation (if any)? 
    \item[] Answer: \answerNA{}
    \item[] Justification: The paper does not involve crowdsourcing nor research with human subjects.

\item {\bf Institutional review board (IRB) approvals or equivalent for research with human subjects}
    \item[] Question: Does the paper describe potential risks incurred by study participants, whether such risks were disclosed to the subjects, and whether Institutional Review Board (IRB) approvals (or an equivalent approval/review based on the requirements of your country or institution) were obtained?
    \item[] Answer: \answerNA{}
    \item[] Justification: The paper does not involve crowdsourcing nor research with human subjects.

\item {\bf Declaration of LLM usage}
    \item[] Question: Does the paper describe the usage of LLMs if it is an important, original, or non-standard component of the core methods in this research? Note that if the LLM is used only for writing, editing, or formatting purposes and does \emph{not} impact the core methodology, scientific rigor, or originality of the research, declaration is not required.
    \item[] Answer: \answerNA{}
    \item[] Justification: The core method development in this research does not involve LLMs as any important, original, or non-standard components.

\end{enumerate}


\end{document}